\documentclass[letterpaper]{article}

\usepackage{times}
\usepackage{helvet} \usepackage{courier}  
\usepackage{graphicx}
\usepackage{amssymb}
\usepackage{amsmath}
\usepackage{amsthm}
\usepackage{bbm}
\usepackage{dsfont}
\usepackage{algorithm}
\usepackage{algpseudocode}
\algrenewcommand\algorithmicrequire{\textbf{Input}}
\algrenewcommand\algorithmicensure{\textbf{Output}}
\DeclareMathOperator*{\argmin}{arg\,min}  
\usepackage{comment}
\usepackage{xcolor}
\usepackage{makecell}
\usepackage{caption,subcaption}
\usepackage[section]{placeins}
\usepackage{url}

\title{An Empirical Study on Ensemble-Based Transfer Learning Bayesian Optimisation with Mixed Variable Types}
\author{Natasha Trinkle\\ RMIT University\\ \and Huong Ha\\ RMIT University\\ \and Jeffrey Chan\\ RMIT University\\ }
\date{January 2026}

\begin{document}

\maketitle

\begin{abstract}
Bayesian optimisation is a sample efficient method for finding a global optimum of expensive black-box objective functions. Historic datasets from related problems can be exploited to help improve performance of Bayesian optimisation by adapting transfer learning methods to various components of the Bayesian optimisation pipeline. In this study we perform an empirical analysis of various ensemble-based transfer learning Bayesian optimisation methods and pipeline components. We expand on previous work in the literature by contributing some specific pipeline components, and three new real-time transfer learning Bayesian optimisation benchmarks. In particular we propose to use a weighting strategy for ensemble surrogate model predictions based on regularised regression with weights constrained to be positive, and a related component for handling the case when transfer learning is not improving Bayesian optimisation performance. We find that in general, two components that help improve transfer learning Bayesian optimisation performance are warm start initialisation and constraining weights used with ensemble surrogate model to be positive.

\end{abstract}

\section{Introduction}
\label{section:introduction}

\subsection{Objectives}

The objective of this work is to provide an empirical analysis of approaches using historic data to improve the performance of Bayesian Optimisation. Bayesian Optimisation (BO) aims to provide a sample efficient approach to optimising black-box objective functions that are typically expensive to evaluate. BO algorithms make use of sequential decision making, active learning and Bayesian decision theory to solve problems where the black-box objective function cannot be optimised using gradient based approaches or assumptions of convexity, but can be evaluated at arbitrary points in the input domain \cite{Kushner1962, Shahriari2016}. While BO uses prior assumptions about objective function characteristics to aid decision making about where to evaluate the expensive objective function for fastest optimisation, it has been suggested that further improvements to the performance of BO may be possible by learning from historic data generated during optimisation of similar processes \cite{Bai2023}. While a number of ways of applying transferring learning to enhance performance of BO (using historic data) have been proposed in literature (\cite{Bai2023}), in this analysis we focus on ways of using information from historic data to initialise the search for a global optimum, and to construct a surrogate model of the black-box objective function, as was explored in \cite{Wistuba2016two, Lindauer2018warmstarting, Feurer2022}. We aim to analyse in detail the pipeline components presented in literature, as well as proposing several additional methods for certain components of this pipeline. We test on a selection of benchmarks from literature as well as three new real-time benchmarks proposed in this work.\\

\subsection{Justification}

Optimisation of expensive black-box functions is a growing research field with applications in a range of different areas. Examples include hyperparameter tuning in automated machine learning models \cite{Shahriari2016, Frazier2018, Brochu2010}, design engineering processes \cite{Frazier2018}, environmental monitoring using expensive to activate sensors \cite{Shahriari2016, Garnett2010}, combinatorial optimisation problems such as the travelling salesman problem (\cite{Jaradat2019, Shahriari2016}), and biological applications such as molecular or drug design \cite{Frazier2018}. More recent examples include machine learning \cite{Bischl2023hyperparameter}, robotics~\cite{Diessner2024development, Li2025b2opt}, chemistry and materials science \cite{Terayama2021black}, environmental monitoring \cite{Hanel2021efficient, Waczak2024physical} and energy management \cite{Ren2023building}. These examples include a number of different types of processes as black-box functions with varying degrees of cost related to evaluating these functions, both in a historic and a real-time context. The different processes contributing to black-box objective functions will also result in different function characteristics. While there are standard sets of prior assumptions commonly used to tackle these problems, sometimes more specific information in the form of historic observations of related problems is available and can also be included in optimisation strategies being explored \cite{Bai2023}. Examples include in AutoML tuning where machine learning algorithm hyperparameters are optimised for many similar datasets \cite{Feurer2022, Bai2023}, as well as applications such as chemical sciences and data-driven experiment planning, where an analytical approach to optimising experimental design based on previous outcomes can significantly reduce experimental effort \cite{Hickman2023equipping, Mahboubi2025point}.\\

\subsection{Background}

 Previous work done on transferring learning from historic datasets to improve performance of BO has focused on four key steps in the BO pipeline as described by \cite{Bai2023}. These include surrogate design \cite{Swersky2013, Poloczek2016,Schilling2016, Wistuba2018scalable, Lindauer2018warmstarting, Feurer2018, Feurer2022}, acquisition function design \cite{Wistuba2018scalable}, initialisation design \cite{Lindauer2018warmstarting} and search space design and pruning \cite{Perrone2019searchspaces}. One approach to surrogate design uses an ensemble of surrogate models consisting of Gaussian Processes, with each one pre-trained on a separate historic task dataset. Motivation for developing an ensemble-based approach to surrogate design in the context of BO with transfer learning was the need to automate the process where by an expert with high levels of experience transfers their knowledge from one problem to another related problem \cite{Bardenet2013collaborative, Wang2024pre} in a way that was computationally tractable \cite{Schilling2016, Wistuba2018scalable}. In particular, there was a need to avoid complex operations on large sets of historic data using multi-task kernels with one Gaussian Process, as had been done in earlier work (see \cite{Swersky2013, Yogatama2014efficient}).\\
 
 Using an ensemble of surrogate models also enables adaptive weighting of historic tasks as new data becomes available \cite{Schilling2016, Wistuba2018scalable}. Weighted predictions from each ensemble surrogate model can be used to construct a surrogate model of the black-box objective function informed by knowledge transferred from these historic related tasks. In the literature there are various ways proposed to combine surrogate model predictions in order to model the black-box objective function \cite{Wistuba2016two, Lindauer2018warmstarting, Feurer2022}. The ensemble has also been used to find warm start initialisation data points instead of random initialisation \cite{Lindauer2018warmstarting}.\\

In this work, we provide an expanded study of previous empirical analyses and ablation studies performed in \cite{Wistuba2016two, Lindauer2018warmstarting, Feurer2022} to provide some suggestions to practitioners looking for best transfer learning with BO strategies to use with particular benchmarks. We propose and test new weighting strategies that show improved performance on some benchmarks, and include new real-time benchmarks with different variable type and dimensions that enable us to demonstrate comparative performance of different approaches on different input domains. In particular, the elements we have added to work done in previous studies include the following: proposal of three new real-time benchmarks to address concerns about OpenML100 datasets (see \url{https://docs.openml.org/benchmark/} \cite{Bischl2017openml, Bischl2019openml}) used to prepare benchmarks proposed in \cite{Perrone2018scalable, Feurer2022}, including one based on OpenML-CC18, addition of weighting strategies based on regularised regression (Lasso and Ridge, with and without a postive constraint on the weights)\footnote{Regularised regression weighting strategies access a continuous weights search space instead of discretised search space as with weighting strategies proposed in \cite{Wistuba2016two, Feurer2022}} and related strategy for selecting between transfer learning and standard BO modes, and inclusion of a set of ranking plots for each benchmark separately that includes all methods explored in this work using warm start initialisation, as well as standard BO\footnote{Ranking plots were included in \cite{Wistuba2016two} for each benchmark separately, where as \cite{Feurer2022} combined all benchmarks into one ranking plot, meaning available plots for analysis of separate benchmarks do not currently include all relevant methods.}. Finally, as part of our empirical analysis, we propose a way to analyse a set of historic datasets with a view to gaining insight about comparative locations of minima points and how this may impact performance of BO with transfer learning based on this these historic datasets.\\

\subsection{Summary of Results}
\label{section:intro_questions}

In this work we explore the following questions:
\begin{enumerate}
    \item How effective are selected ensemble-based transfer learning strategies, including warm start or random initialisation, combined with various weighting strategies in improving performance of Bayesian optimisation as compared to standard Bayesian optimisation (and each other)?
    \item How does inclusion of a strategy for automated selection of standard versus transfer learning Bayesian optimisation (based on measurable criteria available during optimisation) affect the overall performance of a Bayesian optimisation pipeline?
    \item What kind of insights can be obtained from analysing historic datasets about how effective transfer learning with Bayesian optimisation may be on one particular benchmark as compared to another benchmark? 
\end{enumerate}
In addressing these questions we are focussing on three components of the transfer learning BO pipeline presented in previous work \cite{Lindauer2018warmstarting, Wistuba2016two, Feurer2022}. These are initialisation design, computing weights for the surrogate models in the ensemble, and strategies for handling the case where surrogate models are not providing positive transfer learning. By performing a set of ablation studies on various combinations of these components, including those proposed in this work, we seek to empirically analyse and compare the benefits of using these various transfer learning strategies on the selected set of nine benchmarks.\\

In general it is observed that warm start initialisation improves the performance of BO, although for some benchmarks there is variation in results depending on weighting strategy. Comparative performance of different weighting strategies, with both warm start initialisation and random initialisation, appears to be benchmark dependent, although strategies that constrain weights to be positive mostly perform better than those that allow negative weights. Some benchmarks show similar BO performance across most weighting strategies. However other benchmarks exhibit significant BO performance improvement for a particular method(s). The best method(s) is not consistent across benchmarks. For eight out of nine benchmarks, there is at least one method that uses warm start initialisation that outperforms all those using random initialisation, confirming that warm start initialisation makes a significant contribution to improving BO performance using transfer learning. However, impact on BO performance resulting from inclusion of a strategy for automated selection of standard versus transfer learning Bayesian optimisation to handle bad transfer learning is variable, suggesting further work could be done to improve this component of the pipeline. There is some relationship noted between results seen in ranking plots and overlap of location of minima across historic datasets for different benchmarks. However, this relationship across benchmarks is not completely clear, possibly due to impact of dimensionality of input domain and ranking plots including range of weighting strategies with varying performance. Overall, results suggest that, while comparative performance of methods is benchmark dependent, there are general trends around best ensemble-based transfer learning strategies for improving BO performance. These include the use of warm start initialisation, and constraining weights used to combine output from ensemble surrogate models to be positive.\\

\subsection{Guide to Remaining Sections}

The remaining sections of this paper include background information and problem statement in Section \ref{section:background}, a description of methods considered, benchmarks, and experimental details in Section \ref{section:methods}, followed by the main results in Section \ref{section:main_results}. Finally, we conclude with a discussion, including observed limitations to approaches explored, and a conclusion in Section \ref{section:conclusion}.\\


\section{Background \& Problem Statement}
\label{section:background}

To provide context for the work done in this project, a brief background about Bayesian Optimisation and commonly used algorithmic components is provided followed by an introduction to transfer learning.\\

\subsection{Bayesian Optimisation}
\label{subsection:background_related_work_bayesian_optimisation}

Bayesian Optimisation (BO), initially developed by \cite{Kushner1962, Kushner1964new}, was designed to find a global optimum of a stochastic process using Bayes' theorem, where $f$ denotes the model of the stochastic process function and $D_{1,t}$ denotes a dataset of size $t$,  

\begin{equation}
\label{equation:background_bayes_theorem}
    P(f|D_{1:t}) \propto P(D_{1:t}|f)P(f).
\end{equation}

It is an iterative process that requires an initialisation dataset to begin, and then loops through two key steps. Key BO steps include training a surrogate model used to model the black-box objective function given available evaluations, and optimisation of an acquisition function, a utility function that quantifies expected gain in making an evaluation at some input, to find the next input for evaluation. The key aim is to improve on more naive approaches to black-box optimisation including random or grid search in finding an optimal location with a limited evaluation budget \cite{Jones1998efficient}.\\

The first step in BO is to model the objective function using a surrogate model. A Gaussian Process (GP) is a common and suitable choice of surrogate model because it describes a distribution over functions, but also can be used for point-wise predictions since the marginal distribution over a set of evaluations is a multivariate normal distribution \cite{Rasmussen2006, Garnett2023}. A GP, $f(\mathbf{x})$, is fully specified by mean function, $m(\mathbf{x})$, and covariance function, $k(\mathbf{x},\mathbf{x}’)$, 

\begin{equation}
\label{equation:background_related_work_gaussian_process}
    f(\mathbf{x}) \sim \mathcal{GP}\bigl(m(\mathbf{x}),k(\mathbf{x},\mathbf{x}’) \bigl),
\end{equation}

but can also utilise the marginal joint normal distribution over training, and testing outputs to make point-wise predictions, where $y$ denotes $M_{train}$ objective function evaluations (training outputs), $f$ denotes $M_{test}$ test outputs, $\mathcal{K}(X,X)$ is an $M_{train} \times M_{train}$ matrix of covariances computed using the covariance function on the $M_{train}$ training inputs ($X$), $\mathcal{K}(X,X_*)$ is an $M_{train} \times M_{test}$ matrix of covariances on training inputs ($X$), and test inputs $X_*$, $\mathcal{K}(X_*,X)$ is the transpose of $\mathcal{K}(X,X_*)$, $\mathbf{K}(X_*,X_*)$ is the covariance for test inputs, and $\zeta^2I$ is measurement noise,

\begin{equation}
\label{equation:background_related_work_gp_multivariate_normal}
    \begin{bmatrix}y \\ f \end{bmatrix} \sim \mathcal{N}\Biggl(\mathbf{0}, \begin{bmatrix} \mathbf{K}(X,X)+\zeta^2I & \mathbf{K}(X,X_*) \\ \mathbf{K}(X_*,X) & \mathbf{K}(X_*,X_*) \\ \end{bmatrix} \Biggl).
\end{equation}

In order to construct the surrogate model, a dataset of black-box objective function evaluations is required. Therefore, an initial dataset to use for surrogate model construction is needed for the first iteration of the BO evaluation budget. Frazier et al (\cite{Frazier2018}) suggests using a random sampling strategy for this purpose.\\

The second step in BO is to construct a global optimisation policy that enables a trade-off between exploration and exploitation using an acquisition function. There are a number of different acquisition functions used in BO. A common acquisition function is upper confidence bound (UCB), first used in \cite{Lai1985} and applied to GPs by \cite{Srinivas2010},

\begin{equation}
\label{equation:background_related_work_acquisition_function_ucb}
    \alpha_{UCB}(x;\mathcal{D}, \pi) = \mu(x) + \beta\sigma(x)
\end{equation}

where $\pi \in (0,1)$ indicates the confidence level which informs the value of $\beta$ and affects exploration-exploitation trade-off. UCB is simply the mean, $\mu$, plus uncertainty around the mean\footnote{$\mu$ and $\sigma$ are predictions from the surrogate model for some input $x$ \cite{Garnett2023}}. For minimisation problems, lower confidence bound (LCB), $\mu - \beta\sigma$ can be used instead of UCB. \\

In summary, the BO pipeline begins with selection of some initialisation points evaluated on the black-box objective function. This is followed by iterations over construction of a surrogate model (GP) used to make predictions for the black-box objective function, and optimisation of an acquisition function that provides a rationale for selecting the next input point to evaluate on the black-box objective function. The next section (\ref{subsection:transfer_learning}) describes approaches in literature to trying to improve BO performance by transferring learning from historic data generated by related tasks.\\

\subsection{Transfer Learning in Bayesian Optimisation} 
\label{subsection:transfer_learning}

Since many of the optimisation tasks for which BO is used belong to a group of similar tasks, there exist historic datasets from related problems that have been performed previously and can be exploited using transfer learning during a new optimisation task \cite{Bai2023}. For example, when datasets are updated for machine learning tasks, hyperparameters need to be re-optimised \cite{Bai2023}. The term, source tasks, refer to previously performed tasks which are no longer available for evaluation, but from which there exists a historic dataset. The term, target task, refers to the current optimisation task \cite{Pineda2021hpob, Bai2023, Zhuang2020comprehensivetransferlearning}. One challenge in using transfer learning is how to avoid negative transfer. While lacking a rigorous definition, negative transfer could be considered to occur when the contribution made by datasets from previously performed tasks, given the selected algorithm, results in worse performance on the current task than if only data from the current task was included \cite{Wang2019negativetransfer}.\\

One example of using transfer learning in the context of BO is to warm start the search for optima by selecting promising initialisation points based on historic data rather than randomly generating these points as described in Section \ref{subsection:background_related_work_bayesian_optimisation}. In \cite{Feurer2015}, an approach to finding warm start initialisation points based on quantification of relatedness (using metafeatures of task classification datasets) between source and target tasks\footnote{In \cite{Feurer2015}, datasets used by the ML algorithm were compared for this purpose. This is only a feasible approach if one has access to these datasets, which may not always be available.} was proposed. An alternative approach to warm start initialisation was proposed by \cite{Lindauer2018warmstarting} and used by \cite{Feurer2022}. In this approach, an ensemble of surrogate models (one per source task) is constructed. Then all inputs in the combined historic datasets are evaluated on each of the ensemble surrogate models. The input point with best average across all task surrogate models is selected as an initialisation point. Algorithm \ref{algorithm:warm_start} shows how this works to select multiple initialisation points.\\

Another example of transfer learning in the context of BO is ensemble learning. In particular, Schilling et al (\cite{Schilling2016}) proposed the use of an ensemble of experts  consisting of GPs pretrained on historic data from a set of source tasks. As target observations become available, these GPs were further trained on the target dataset. Predictions were made using this ensemble of GPs which is the combination of predictions of each GP expert. Simple weighting schemes were used that ranked all source tasks equally. In another work, \cite{Wistuba2016two} proposed a Two-Stage Transfer Surrogate model using pairwise ranking of inputs for weighting (TSTR). This also involved the construction of an ensemble of GP surrogate models, one per historic task, in the first of the two stages. Their weighting scheme involved using Epanechnikov quadratic kernel to compute distance between two datasets, combined with Nadaraya Watson kernel to compute predicted means. In this work \cite{Wistuba2016two} used performance rankings\footnote{Performance ranking loss function uses number of discordant pairs between two datasets \cite{Wistuba2016two}.} and the metric to compare hyperparameter datasets.\\

A different approach was explored by \cite{Lindauer2018warmstarting}. This approach permitted negative weights, and used stochastic gradient descent on a mean squared error loss function to construct the target model as a weighted linear combination of predictions from the ensemble models. Training and validation sets were employed to avoid over-fitting. The work of \cite{Feurer2018, Feurer2022} proposed a Ranking-Weighted Gaussian Process Ensemble (RGPE) and used a linear combination of predictions from ensemble surrogate models with weighting scheme adapted from that proposed by \cite{Lacoste2014} and extending the approach using discordant pairs proposed by \cite{Wistuba2016two}.\\

The next section contains a formal statement of the problem with mathematical definitions of the problem space and historic data used for transfer learning. 

\subsection{Regularised Regression}
\label{section:regression}

In the previous section (\ref{subsection:transfer_learning}), there are several weighting schemes described that were used to linearly combine output from the set of surrogate models in an ensemble to produce a prediction on the target problem. Another approach to weighting ensemble surrogate models would be to use regularised regression. Regularised regression involves using mean squared error as a loss function combined with some term to prevent over-fitting of the weights. For an in depth description, please see \cite[Chapter 3]{Hastie2009elements}. Regularised regression is useful in the case of ill-posed problems where the inverse problem does not necessarily have an exact solution \cite{Tikhonov1995numerical,Wahba2019representer}. Two popular forms of regularisation are Ridge, which minimises a loss function with an L2 penalty or shrinkage term \cite{Hastie2009elements}, and Lasso, proposed by \cite{Tibshirani1996regression}, which uses an L1 penalty term. The L1 penalty term results in some weights shrinking to $0$, resulting in simpler models (see \cite{Tibshirani1996regression} for more details). Equation \ref{equation:background_ridge_loss} shows the loss function for Ridge regularised regression, and Equation \ref{equation:background_lasso_loss} shows the loss function for Lasso regularised regression. In both equations, $w_i$ represents the weights used on the $i \in 1,...,N$ features, which are represented as $h_{i,j}$ where $j \in 1,...,M$ represents each of the input points, $y_{target,j}$ represents the true function output for each of the $M$ input points, and $\alpha$ represents the penalty hyperparameter.\\

\begin{equation}
    \label{equation:background_ridge_loss}
    \mathcal{L}_{Ridge} = \frac{1}{M} \sum_{j=1}^{M}(y_{true,j} - \sum_{i=1}^{N}w_ih_{i,j})^2 + \alpha\sum_{i=1}^{N}w_i^2
\end{equation}\\

\begin{equation}
    \label{equation:background_lasso_loss}
    \mathcal{L}_{Lasso} = \frac{1}{M} \sum_{j=1}^{M}(y_{true,j} - \sum_{i=1}^{N}w_ih_{i,j})^2 + \alpha\sum_{i=1}^{N}|w_i|
\end{equation}\\

The next section provides a formal description of the problem being addressed using BO in this work, including the use of historic datasets for transfer learning.\\


\subsection{Problem Statement}
\label{section:problem_statement}

A single-objective target function is optimised over a mixed-variable search space, described as $\phi_{target}: \mathcal{H} \times \mathcal{Z} \times \mathcal{C} \rightarrow \mathbb{R}$, where $\mathcal{H} = \mathcal{H}^{(1)} \times...\times \mathcal{H}^{(d)}$ is the domain of categorical parameters, $\mathcal{Z} = \mathcal{Z}^{(1)} \times...\times \mathcal{Z}^{(d_z)}$ is the domain of discrete (integer) parameters, and $\mathcal{C} = \mathcal{C}^{(1)} \times...\times \mathcal{C}^{(d)}$ is the domain of continuous parameters. The number of categorical parameters is $d_h\geq 0$, the number of discrete parameters is $d_z\geq 0$, and the number of continuous parameters is $d_c\geq 1$. Let $\mathbf{x} = [\mathbf{h}, \mathbf{z}, \mathbf{c}]$, where $\mathbf{h} = [h_1,h_2,...,h_{d_h}]$, $\mathbf{z} = [z_1,z_2,...,z_{d_z}]$ and $\mathbf{c} = [c_1,c_2,...,c_{d_c}]$ and $\mathcal{X} = \mathcal{H}^{(1)} \times...\times \mathcal{H}^{(d_h)} \times \mathcal{Z}^{(1)} \times...\times \mathcal{Z}^{(d_z)} \times \mathcal{C}^{(1)} \times...\times \mathcal{C}^{(d_c)}$ \cite{Daulton2022}. The optimisation (minimisation) task can be expressed as  

\begin{equation}
\label{eqn:e3}
\mathbf{x}^* = \mathop{\arg \min}_{\mathbf{x} \in \mathcal{X}} \phi_{target}(\mathbf{x}).
\end{equation}\\

In the context of BO with transfer learning, it is assumed that, prior to any evaluations being made on the target function, there exists a set of $N$ historic datasets which were generated previously (ie. during optimisation) on $N$ related source tasks\footnote{Note that historic datasets, generated during some optimisation process, will not share common input configurations.} as in \cite{Feurer2022}. Source tasks are related to the target task via a shared input domain resulting from some similar process. Source task objective functions are no longer available for evaluation during target task optimisation. Each of the $N$ datasets containing $M_i$ noisy evaluations can be expressed, $\mathcal{D}_{source} = \{\mathcal{D}_1,...,\mathcal{D}_N\}$, where $\mathcal{D}_i = \{\boldsymbol{x}_{ij}, y_{ij}\}_{j=1}^{M_i}$.\\

It is also assumed that all source and target task objective functions can be approximated using a surrogate model (GP). For task $i$, $f_i \sim \mathcal{GP}(\mu_i, K_i)$, where $\mu_i$ indicates the mean function of $f_i$, and $K_i$ is the kernel (covariance) function. We use an ensemble of surrogate models, inclusive of all source task GPs and a GP to model the target task, $f_{target}$, and we denote the ensemble of surrogate models as $f_{ens} = \{f_i\}_{i=1}^N \cup f_{target}$. \\

\section{Methods}
\label{section:methods}

The methods we are presenting for consideration here are broken down into components of the overall pipeline used for BO with ensemble transfer learning as presented in literature \cite{Lindauer2018warmstarting, Wistuba2016two, Feurer2018, Feurer2022}. We also propose four new weighting strategies in Section \ref{section:weights}, and one new strategy for handling bad transfer learning in Section \ref{section:methods_handling_worst_case}.

\subsection{Pipeline Overview}
The general pipeline can be seen in Algorithm \ref{algorithm:overall_pipeline}. In this work, we consider methods for initialisation of the target dataset (Section \ref{section:initialisation}), construction of the ensemble of surrogate models, and in particular the weighting strategies used (Section \ref{section:weights}), and finally, approaches to reverting to standard BO when transfer learning is not improving BO performance (Section \ref{section:methods_handling_worst_case}).\\

\begin{algorithm}
\caption{Bayesian Optimisation Transfer Learning Pipeline}
\label{algorithm:overall_pipeline}
\begin{algorithmic}[1]
    \Require 
    $\mathcal{D}_{source}= \{\mathcal{D}_1,...,\mathcal{D}_N\}$ (historic data), $\mathcal{T}$ (eval budget), $n\_init$ (number of initialisation points)
    \State \textbf{Initialisation} 
        \State $f_{ens} \gets \Call{ConstructEnsemble}{\mathcal{D}_{source}}$
        \State $\{\boldsymbol{x}_{init_j}\}_{j = 1}^{n\_init} \gets \Call{WarmStart}{f_{ens}}$ (see Algorithm \ref{algorithm:warm_start})
        \State $\{y_{init_j}\}_{j = 1}^{n\_init} = \phi(\{\boldsymbol{x}_{init_j}\}_{j = 1}^{n\_init})+\epsilon_{target}$ (objective function evaluation)
        \State $\mathcal{D}_{target} \gets \{\boldsymbol{x}_{init_j}, y_{init_j}\}_{j = 1}^{n\_init}$
        \State Specific method initialisation procedures (see Section \ref{section:weights})
    \For{$i=n\_init$ to $\mathcal{T}$}
        \State \Call{UpdateEnsemble}{$\mathcal{D}_{target}$, $f_{ens}$}
        \State \Call{ComputeEnsembleWeights}{$\mathcal{D}_{target}$, $f_{ens}$} (see Section \ref{section:weights})
        \State $x_{target_i} = \argmin_{x \in \mathcal{X}} LCB (x)$ (see acquisition function optimisation, Section \ref{subsection:background_related_work_bayesian_optimisation})
        \State $y_{target_i} = \phi(x_{target_i}) + \epsilon_{target}$ (objective function evaluation)
        \State $\mathcal{D}_{target} = \mathcal{D}_{target} \cup \{{x}_{target_i}, y_{target_i}\}$
    \EndFor
    \Ensure $\mathcal{D}_{target}$, $x_{target_{best}}$
    \Statex
    \Function{ConstructEnsemble}{$\{\mathcal{D}_i\}_{i=1}^N$}
        \State Initialise $f_{ens}$
        \For{$i=1,..,N$}
            \State Train GP on $\mathcal{D}_i$ ($\{f_i \sim \mathcal{GP}(\mu_i, K_i)\}_{i=1}^N$)
            \State Append GP to $f_{ens}$
        \EndFor
        \State \Return $f_{ens}$
    \EndFunction
    \Statex    
    \Function{UpdateEnsemble}{$\mathcal{D}_{target}$,$f_{ens}$}
        \If{$f_{target} \in f_{ens}$}
            \State Remove $f_{target}$
        \EndIf  
        \State Train GP on $\mathcal{D}_{target}$ ($f_{target} \sim \mathcal{GP}(\mu_{target}, K_{target})$)
        \State Append target GP to $f_{ens}$
        \State \Return $f_{ens}$
    \EndFunction
        
\end{algorithmic}
\end{algorithm}

Equation \ref{equation:ensemble}, where $f_{ens}(x_j)$ is the output normal distribution for input $x_j$, $w_i$ denotes weights on source surrogate model mean predictions, $w_{target}$ is the weight on target surrogate model mean prediction, and variance of $f_{ens}(x_j)$, denoted $\sigma_{target}^2(x_j)$, is obtained only from the target surrogate model,

\begin{equation}
\label{equation:ensemble}
f_{ens}(x_j) \sim \mathcal{N}\big(\sum_{i=1}^{N}w_i\mu_i(x_j) + w_{target}\mu_{target}(x_j), \sigma_{target}^2(x_j)\big),
\end{equation}

describes the construction of the ensemble of GP surrogate models evaluated at a particular input point as a normal distribution. Unlike the standard BO (no transfer learning) setting, where the target objective function is modeled using a single GP constructed from the current observed dataset, in this transfer learning based BO approach, the target objective function is modeled using an ensemble of GPs that are constructed from the historic datasets to transfer information in the historic datasets to the new target problem. Note that a surrogate model of the target objective function is included, and updated at each iteration. Inclusion of the target surrogate model helps to model the target objective function in cases where the source surrogate models are inadequate \cite{Wistuba2016two, Feurer2022}. \\

\subsection{Initialisation}
\label{section:initialisation}
For the initialisation step (Lines 1-6 in Algorithm \ref{algorithm:overall_pipeline}) we compare two approaches to generating initialisation points. The first of these is random generation of a set of $n\_init$ initialisation points using Latin hypercube sampling. The second uses historic datasets to find a set of $n\_init\_ws$ warm start initialisation points \cite{Lindauer2018warmstarting} (see Algorithm \ref{algorithm:warm_start} for details). Both approaches use the implementation from \cite{Feurer2022}. We compare how choosing one or other of these initialisation approaches described impacts performance of standard BO (see Section \ref{section:exp_setup} for setup details), and then the seven weighting schemes described in Section \ref{section:weights}.\\

\begin{algorithm}
\caption{Warm Start Initialisation}
\label{algorithm:warm_start}
\begin{algorithmic}[1]
    \Require $X_{candidates}=\{\cup \{\boldsymbol{x}_{i,j}\}_{j = 1}^{M_i}\}_{i=1}^{N}$ (union of historic dataset (inputs)), $n\_init\_ws$ (number of required initialisation points), $f_{ens}$ (pre-trained ensemble)
    \State $X_{inits} \gets \text{empty list}$
    \For{$p=1$ to $n\_init\_ws$}
        \Statex{$MeanBestEvals \gets \text{empty dictionary}$}
        \For{$x$ in $X_{candidates}$}
            \State{$BestEvals \gets \text{empty list}$}
            \For{$q=1$ to $N$}
                \State{$BestEvals[q] \gets \min(\mu_q(x), \{\mu_k(x)\}_{k=1}^{p-1})$} 
            \EndFor
            \State{$MeanBestEvals[x] \gets \frac{1}{N}\sum_{q=1}^N(BestEvals[q])$}
        \EndFor
        \State Append $x$ to $X_{inits}$ which corresponds to min val in $MeanBestEvals$
    \EndFor
    \Ensure $X_{inits}$
\end{algorithmic}
\end{algorithm}    

\subsection{Weighting Strategies for the Ensemble of Surrogate Models}
\label{section:weights}

We consider seven different weighting strategies to find $\{w_i\}_{i=1}^N$ and $w_{target}$ for use with the ensemble of surrogate models (Equation \ref{equation:ensemble}). The first four weighting strategies involve regularised regression, which formulates the observed objective function values, $y_{target}$, to be equal to a weighted linear combination of predictions from the GPs in the ensemble. Regularised linear regression is used to find the appropriate contributing weight of each GP by minimising a loss function that includes a penalty on the weights as described in Section \ref{section:regression}, and Equations \ref{equation:background_ridge_loss} (for Ridge) and \ref{equation:background_lasso_loss} (for Lasso). In literature, \cite{Lindauer2018warmstarting} uses stochastic gradient descent of the mean squared error loss function with L2 regularisation (equivalent to Ridge regularised regression). We extend this approach, and include not only Ridge regularised regression with an L2 penalty for its suitability to handle ill-posed problems \cite{Tikhonov1995numerical,Wahba2019representer}, but also Lasso with an L1 penalty for its ability to reduce model complexity by shrinking some weights to $0$ \cite{Tibshirani1996regression}. We also include each of these two methods with a constraint on the weights, forcing them to be positive. The intuition behind constraining weights to be positive in the context of transfer learning is that we are using functions from historic tasks we believe to be similar to the target task. In the context of using regression with basis functions in the hypothesis space, we want to be able to positively or negatively weight the basis functions. However, in this context where we use related or similar functions in the hypothesis space, it is unlikely that the reflection of any function ($\phi(x)*(-1)$) will result in positive transfer of information to the target function. Therefore, constraining weights to be positive helps to avoid negative transfer of information (see Section \ref{subsection:transfer_learning}). Equations \ref{equation:method_ridge_loss} and \ref{equation:method_lasso_loss},

\begin{align}
\label{equation:method_ridge_loss}
    \text{minimise}\qquad & \frac{1}{M_{target}}\sum_{j=1}^{M_{target}}(y_{target}(x_{j}) - \sum_{i=1}^{N+1}w_i\mu_{i}(x_{j}))^2 + \alpha\sum_{i=1}^{N+1}w_i^2\\ \notag
    \text{subject to }\qquad & w_i >=0 \qquad \forall i \in \{1,...,N+1\}\\ \notag
\end{align}

\begin{align}
\label{equation:method_lasso_loss}
    \text{minimise}\qquad & \frac{1}{M_{target}}\sum_{j=1}^{M_{target}}(y_{target}(x_{j}) - \sum_{i=1}^{N+1}w_i\mu_{i}(x_{j}))^2 + \alpha\sum_{i=1}^{N+1}|w_i|\\ \notag
    \text{subject to }\qquad & w_i >=0 \qquad \forall i \in \{1,...,N+1\}\\ \notag
\end{align}

describe the weighting loss function to be minimised for Ridge and Lasso when weights are constrained to be positive, where $w_i \in 1,...,N+1$ represent the $N+1$ weights on all $N$ source surrogate models plus $1$ target surrogate model in the ensemble, $\mu_{i}(x_{j})$ represents the predicted mean value from the $ith$ surrogate model on the $jth$ input point ($j \in 1,...,M_{target}$ where there are $M_{target}$ input points), $y_{target}(x_{j})$ is the target objective function true evaluation, and $\alpha$ represents the penalty hyperparameter. Also, rather than just computing weightings on training data, we use bootstrap with $1000$ samples. We use the abbreviation LaGPE for the Lasso Gaussian Process ensemble methods, and RiGPE for Ridge Gaussian Process ensemble methods. In summary, we test four different regularised regression methods, LaGPE with a positive constraint on the weights, LaGPE without a positive constraint on the weights, RiGPE with a positive constraint on the weights and RiGPE without a positive constraint on the weights, to evaluate the effectiveness of the positive constraint.\\

The remaining three weighting strategies tested have been extracted from approaches proposed in literature. The first of these is a ranking approach, Ranking-Weighted Gaussian Process Ensemble (RGPE) proposed by \cite{Feurer2022}. In this approach, weights are computed by ranking all source functions based on number of discordant pairs as compared to the target objective function. For all possible pairings of input points in the training data, the number of pairings where the surrogate mean prediction function increase or decrease is different to the objective target function are summed. Bootstrap with sampling of $1000$ is used to compute a distribution over the number of times a particular source surrogate model appears in the group of source models with minimum number of discordant pairings. Equation \ref{equation:ranking_loss},

\begin{equation}
\label{equation:ranking_loss}
    \mathcal{L}_{i,s}(f_{i}, \mathcal{D}_{target}) = \sum_{j=1}^{M_{target}}\sum_{k=1}^{M_{target}}\mathds{1}((f_{i}(x_{target_j}) < f_{i}(x_{target_k}))\oplus(y_{target_j} < y_{target_k})),
\end{equation}

shows the loss function used for the number of mis-ranked pairs between bootstrap sample $s$ for the $ith$ source task surrogate model predicted mean function and the target objective function. Equation \ref{equation:rgpe_weight_calc},

\begin{equation}
\label{equation:rgpe_weight_calc}
    w_i = \frac{1}{1000}\sum_{s=1}^{1000} \biggl(\frac{\mathbb{I}(i \in \argmin \{\mathcal{L}_{j,s}\}_{j=1}^{N+1})}{\sum_{k=1}^{N+1}(k \in  \argmin \{\mathcal{L}_{k,s}\}_{k=1}^{N+1})} \biggl),
\end{equation}

shows how weights are computed using RGPE. See \cite{Feurer2022} for more details. \\

The second of these is the Two-Stage Transfer Surrogate Model with Rankings (TSTR) method from \cite{Wistuba2016two}, which uses the Nadaraya-Watson kernel weighted average to compute weights in the surrogate models in the ensemble as seen in Equation \ref{equation:nadaraya}

\begin{equation}
\label{equation:nadaraya}
    w_i = \frac{k_p(\mathcal{X}_i, \mathcal{X}_{target})}{\sum_{j=1}^{N+1}k_p(\mathcal{X}_i, \mathcal{X}_{target})}.
\end{equation}

The $\mathcal{X}$ here indicates the datasets constructed from ensemble surrogate model mean predictions on the target dataset inputs or target dataset. Equation \ref{equation:epan},

\begin{align}
\label{equation:epan}
    k_p(\mathcal{X}_i, \mathcal{X}_{j}) = \gamma\biggl(\frac{||\mathcal{X}_i - \mathcal{X}_{j}||_2}{\rho}\biggl)\\ \notag
    \gamma(r) =     
    \begin{cases}
      \frac{3}{4}(1-r^2) & \text{if $r \leq 1$}\\
      0 & \text{otherwise}\\
    \end{cases}  
\end{align}

is the exact Epanechnikov quadratic kernel used to compute distance between the surrogate model mean prediction evaluations and the target objective function evaluations. Again the number of discordant pairs is computed, but this time as the distance between the datasets $\mathcal{X}_i$ and $\mathcal{X}_{target}$ used in kernel. Here $\rho$ is a bandwidth hyperparameter that needs to be set by the user.\\

The third literature weighting strategy, from \cite{Lindauer2018warmstarting}, is the warm-starting algorithm configuration (WAC). Similar to our regularised regression weighting strategies, WAC instead uses stochastic gradient descent to minimise a mean squared error loss function with L2 regularisation to weight surrogate models in the ensemble. In contrast to all other methods, this method uses prediction uncertainty from all surrogate models in the ensemble in making a prediction\footnote{Note that only the predicted mean values are used for training the weights. These weights are then used to compute a linear combination of predicted mean and variance for prediction (see GitHub repository from \cite{Feurer2022}).}, rather than only the target GP uncertainty.\\

Additionally, each weighting strategy has different requirements in terms of hyperparameter tuning and for early iterations. One of the challenges in using a Ridge or Lasso loss function is tuning the penalty coefficient, $\alpha$. In this work, cross validation is used to learn the best $\alpha$. The value of $\alpha$ is important in selecting which variables are correlated to the target function and contribute to predictions made on the target function \cite{Su2017false}. However, in the context of BO, the dataset size is so small during early iterations that measurement of cross validation error\footnote{In sklearn LassoCV, used in this work, validation set error is computed using $R^2$, which requires at least two validation points. If the number of splits is 3, at least 6 data points are required to use cross validation to find best $\alpha$ \cite{Scikit-learn}.} is unreliable. This was observed to lead to overly large estimations of $\alpha$ during early iterations resulting in a weighting of $0$ on all source surrogate models. To address this problem, we pre-learn a reasonable guess for $\alpha$ for the target task by using the historic datasets in the following way. Iterating through all historic task datasets, we assign one task, $i$, as the pseudo-target task. Then we compute best $\alpha$ for the historic dataset $\mathcal{D}_i$, of that task using an ensemble of surrogate models training on remaining historic datasets, $\mathcal{D}_{source} \setminus \mathcal{D}_i$. We use Ridge or Lasso regression with $\mathcal{D}_i$ as the training data and predictions from the ensemble as features, and find best $\alpha$ for historic task $i$ using cross validation. After iterating through all historic tasks, we take the median of the best $\alpha$s computed for all historic task datasets. See Algorithm \ref{algorithm:alpha} in Appendix \ref{section:alpha_alg} for details.\\

For the RGPE weighting strategy, there are no hyperparameter requiring tuning. However, in the implementation in \cite{Feurer2022}, for the third iteration of the evaluation budget weights of $\frac{1}{N}$ are used for all models in the ensemble. For TSTR the bandwidth hyperparameter requires tuning. For convenience, we pick one value, $0.1$, used as a default in the \cite{Feurer2022} implementation of this method, and use that for all benchmarks. Consequently, it is possible that TSTR may perform better if more tuning of bandwidth were undertaken. Finally, no special tuning or iterations were required for the WAC method.\\

\subsection{Strategy for Handling Bad Transfer Learning}
\label{section:methods_handling_worst_case}

This component in the pipeline is designed to prevent the problem of wasting expensive evaluations exploring unfruitful regions of the input domain due to misguidance from source datasets and surrogate models. We look at two different design approaches for this component on a subset of methods described in the weighting strategies section (\ref{section:weights}). The first, a method designed to prevent "weight dilution" of source surrogate models, was implemented in \cite{Feurer2022} and is applied here to the RGPE and TSTR methods\footnote{In the implementation of these two methods provided by \cite{Feurer2022}, this weight dilution prevention strategy is part of the pipeline. WAC is excluded from this analysis as in the implementation provided this component was not included for this method.}. It involves computing a probability of dropping a particular model from the ensemble based on how it compares to the target surrogate model in the ensemble (comparing number of discordant pairs with the target objective function). Equation \ref{equation:pdrop} provides details, 

\begin{equation}
\label{equation:pdrop}
    p_{drop_i} = 1 - \biggl((1-\frac{t}{\mathcal{T}})\frac{\sum_{s=1}^{S}\mathds{1}(\mathcal{L}_{i,s}<\mathcal{L}_{target,s})}{S}\biggl),
\end{equation}

where $\mathcal{T}$ indicated the evaluation budget, $t$ the current iteration number, $S$ the number of bootstrap samples, and $\mathcal{L}$ the number of discordant pairs for surrogate model $i$ and bootstrap sample $s$. A key feature of this method is that the probability is tied to iteration number such that when the evaluation budget is exhausted, the probability of dropping a particular source surrogate model will be $1$ if number of discordant pairs is greater than for the target GP.\\

The second method, contributed as part of this work, is based on the mean squared error loss, and is applied to the LaGPE and RiGPE weighting strategy methods. It has been designed to not require an additional hyperparameter to tune. In extending the weighting strategies to include handling poor transfer learning, we assume that optimisation can occur via two modes. There is mode one, using the transfer learning ensemble, or mode two using only the target GP. In this approach the average mean squared error computed during cross validation to find best regularisation hyperparameter is used to track quality of hypothesis space (ensemble surrogate models or target GP) in modelling the target objective function. The average mean squared error from the previous two iterations is converted to a probability of changing from current mode to the other mode, and multiplied by an indicator of model performance; $0$ if previous iteration yielded the optimal evaluation so far and $1$ if not. Hence, while the model is actively optimising, probability of changing mode will be $0$. Equation \ref{equation:prob_change}, 

\begin{align}
\label{equation:prob_change}
    p_{change} = \mathbb{I}(y_{target_{t-1}}>\min \{y_{target_k}\}_{k=1}^{t-2})*\biggl (\max(0, \frac{(mse_{t-1}-\frac{(mse_{t-1}+mse_{t-2})}{2})}{\frac{(mse_{t-1}+mse_{t-2})}{2}}  \biggl) \\ \notag
    \text{where} \\ \notag
    mse = 
    \frac{1}{K}\sum_{k=1}^{K}\biggl(\frac{1}{M_{target}/K}\sum_{j=1}^{M_{target}/K}(y_{target}(x_{j}) - \sum_{i=1}^{N+1}w_i\mu_{i}(x_{j}))^2 \biggl) \\ \notag
\end{align}

is used to compute this probability, where $\mathbb{I}$ represents the indicator function for model performance, and $mse$ is the average computed mean squared error between objective function evaluations and the weighted linear combination of surrogate model mean predictions for each split, in $K$-fold cross validation at each iteration.\\

One of the key differences between the two approaches, apart from using a different loss metric, is that the weight dilution prevention strategy in \cite{Feurer2022} selectively eliminates source surrogate models individually, where as we propose a more abrupt method where we use all transfer learning BO or all standard BO. We leave a more in depth comparative analysis of these two approaches to another time.\\

\section{Experimental Setup}
\label{section:exp_setup}

This section provides details of experimental set up used by all methods. Then we provide details about the benchmarks we have selected. Finally we discuss metrics we use for evaluation.\\

\subsection{Setup for the Gaussian Process}
\label{section:standard_bo}

All weighting methods tested in this study use an ensemble of GP models. A GP is also the chosen surrogate model for standard BO methods described in Section \ref{section:methods}. All GP models used in this study are set up with identical prior mean and kernel functions. We use a prior mean function of zero. The kernel used is a combination of a Matérn kernel with $\nu=2.5$ for continuous and integer variables, and a Hamming kernel for categorical variables. Independent, identically distributed Gaussian noise is modelled using a white noise kernel. This implementation follows those used in \cite{Lindauer-jmlr22a, Feurer2022}.\\

\subsection{Other Details}
For all methods use an an evaluation budget of $100$ to see effect of certain choices more clearly than in work done previously (\cite{Feurer2022}). The historic datasets are constructed from $50$ evaluations from the standard BO run in keeping with \cite{Feurer2022} implementation. For initialisation of the target dataset, we either use $10$ randomly generated initialisaiton points (Latin hypercube sampling), or $2$ warm start initialisation points (using historic datasets) \cite{Lindauer2018warmstarting}. We use LCB as the acquisition function because it is simple to compute and a popular acquisition function in the BO literature (see Section \ref{subsection:background_related_work_bayesian_optimisation} for details). \\

The BO package we use for this work is the SMAC package \cite{Lindauer-jmlr22a} to enable fair comparison with baseline implementations from \cite{Feurer2022}, and SMAC settings were carefully selected to be consistent with these implementations. All methods use the same kernel in single task GPs and have been adapted to use LCB acquisition function. \\

\subsection{Benchmarks}
\label{section:benchmarks}

In this work, we focus on parameter and hyperparameter tuning of a selection of machine learning, regression and simulation black-box objective functions that can be either easily evaluated, or modelled using a surrogate model to provide approximate evaluations in a reasonable time frame. Suitable benchmarks for transfer learning in a BO context require a set of contextually related tasks with same input domain that each mimic an expensive black-box objective function as described in Section \ref{section:problem_statement}, while still being feasible in terms of runtime and memory to use for development and testing \cite{Eggensperger2015efficient}. In the experimental set up proposed in \cite{Feurer2022} and also used in this work, each benchmark consists of $N+1$ tasks, such that $1$ task can be selected as the target task, leaving the remaining $N$ as source tasks. We run each method for each of the $N+1$ tasks acting as target task on 15 seeds. Results are averaged over $(N+1)*15$ runs for one method. This approach requires there to be no hierarchy or dependency amongst tasks. We use four different types of benchmarks for validation that meet these criteria. These include grid, surrogate, simulation, and real-time time series benchmarks.\\

The grid benchmark, a neural network benchmark, contains 4 continuous variables and 3 integer variables\footnote{For the grid benchmark, only a dataset of 2000 data points per task is available. This is used in place of the objective function. Therefore, recommended evaluation configurations must be match to their nearest data point using a strategy like nearest neighbours.}. It was designed for multi-fidelity optimisation by \cite{Zimmer2021} and adapted to the transfer learning scenario by \cite{Feurer2022}. It consists of 35 datasets, with each dataset containing 2000 randomly sampled hyperparameter configurations from funnel-shaped MLP nets \cite{Gijsbers2019open}. For details about variables used in this benchmark please see \cite[Appendix D, Table 14]{Feurer2022} and accompanying code.\\

The five surrogate benchmarks consist of between 2 and 10 variables, with between 2 and 10 variables being continuous, between 0 and 5 variables being integer (discrete), and up to 2 variables being categorical. Each surrogate benchmark consists of a set of 38 configuration datasets generated by running five different supervised machine learning algorithms on 38 classification tasks with 2000 different hyperparameter configurations. The five machine learning algorithms include svm, glmnet, rpart, ranger and xgboost. For openml-glmnet there are 2 continuous variables. For openml-svm there are 1 categorical variable, 1 continuous variable, and 1 integer variable that are conditional on the categorical variable, and 1 additional continuous variable. For openml-rpart there is 1 continuous variable and 3 integer variables. For openml-ranger there are 3 continuous variables, 2 categorical variables and 1 integer variable. For openml-xgb (xgboost) there are 10 variables with 1 categorical variable, 3 continuous and 1 integer variable conditional upon the categorical variable, plus 1 integer variable and 4 continuous variables\footnote{The two benchmarks, openml-svm and openml-xgb, which include conditional hyperparameters, have a set of active hyperparameters that are conditional on the value of a categorical hyperparameter. For example, openml-svm has a categorical variable for kernel type which can take values; 'linear', 'polynomial', 'radial'. For 'linear', no additional variable is required, for 'polynomial' a degree variable is required, and for 'radial' a gamma variable is required. The SMAC package used for this work handles these natively \cite{Lindauer-jmlr22a} according to approach described in \cite{Levesque2017bayesian}.}. The objective value selected to be used for these benchmarks is auc\footnote{Performance measures recorded by \cite{Kuhn2018automatic} included auc, accuracy, brier score and runtime. However, auc was used in the implementation by \cite{Feurer2022}.}. The preparation of the surrogate benchmarks from these datasets involved training a Random Forest regression model on the available dataset so that it could be used to approximate the evaluations of the benchmark algorithm used to generate the dataset. Methodology for surrogate model construction implemented by \cite{Feurer2022} was from \cite{Eggensperger2015efficient}\footnote{Two of the surrogate benchmarks described here, openml-rpart and openml-ranger, did not appear in the publication \cite{Feurer2022}. However, they are part of the datasets prepared in \cite{Kuhn2018automatic}, and were included with methodology for use as a RandomForest surrogate benchmark implementation in the \cite{Feurer2022} GitHub repository }.\\

The 38 classification tasks used to generate the 38 configuration datasets for each of the machine learning algorithms were selected from a larger set of 100 classification datasets available as the OpenML-100 benchmark suite \cite{Vanschoren2014openml, Bischl2017openml}. The subset of 38 datasets were selected to avoid missing values and because they have binary output. Originally proposed and used as a benchmark for BO with transfer learning by \cite{Perrone2018scalable}, the configuration datasets for these benchmarks implemented by \cite{Feurer2022}, was prepared by \cite{Kuhn2018automatic}. Please note that since the proposal of these benchmarks using the OpenML-100 benchmark suite, the OpenML website (\url{https://docs.openml.org/benchmark/}) now includes a comment recommending using the OpenML-CC18 benchmark suite rather than OpenML 100 which "suffers from some teething issues" and "may obfuscate interpretation of results". For details about variables used in Openml100 benchmarks implemented by \cite{Feurer2022}, please see Appendix \ref{appendix:openml100_details}.\\

As a result of OpenML recommendation to use the OpenML-CC18 benchmark suite, we propose a benchmark using the RandomForest machine learning algorithm \cite{Scikit-learn} on a subset of OpenML-CC18 datasets \cite{Bischl2019openml}. This $10$ dimensional benchmark consists of 3 continuous variables, 5 integer variables and 2 categorical variables. A random selection of 38 classification datasets was selected from OpenML-CC18 based on practical considerations around runtime. Input domain for each variable was selected using trial and error to ensure reasonable performance across all tasks. The evaluation metric used as black-box objective function output is accuracy. This metric was chosen to as it is appropriate for classification datasets with differing number of classes and class imbalance. For a list of OpenML-CC18 task IDs please see Appendix \ref{appendix:randomforest_details}, and for more details about variables used please see Table \ref{tab:randomforest_bench}.\\

We also propose a benchmark constructed using LassoBench contributed by \cite{Sehic2022}. This benchmark consists of a version of Lasso regression model that requires one regularisation hyperparameter per feature in the regression dataset. In this work we use the implementation provided by \cite{Sehic2022}, but adapt it to our own BO setup, and use data from another source appropriate to our transfer learning experimental requirements. The data we selected for regression is from the California Cooperative Oceanic Fisheries Investigations website (\cite{CalCOFI2025}) and consists of a truncated set of features from the Bottle Database. These $10$ features include characteristics of sea-water collected including depth, salinity, oxygenation and other nutrients. The predicted variable is water temperature. This results in a benchmark with $10$ tunable continuous hyperparameters for datasets with $10$ features used to model the output variable in regression. This dataset was selected because it can be split into $59$ different tasks consisting of different locations off the coast of California from which the sea-water samples were collected. The metric used as black-box objective function evaluation output is validation mean squared error. For more details about variables used, please see Table \ref{tab:lassobench_bench} in Appendix \ref{appendix:lassobench_details}.\\

The simulation benchmark, based on linear quadratic regulator (LQR) control of a cartpole, was designed especially for this study to provide an alternative to machine learning benchmarks. It is a $2$ dimensional continuous variable benchmark, and the setup is adapted from a a multi-task learning scenario \cite{Marco2017virtual}, to the transfer learning with BO context being studied in this work. The black-box objective function evaluation output is a cost function depending on the cart position, the pole angle, the voltage, summed over all the time steps of one simulation \cite{Marco2017virtual} (see Equation \ref{equation:cartpole_cost}). Please see Appendix \ref{appendix:cartpole_details} for more details and Table \ref{tab:cartpole_bench} for variable descriptions.\\

\subsection{Metrics}

In order to evaluate and compare performance of BO with transfer learning using an ensemble of surrogate models on the nine benchmarks described in Section \ref{section:benchmarks}, we select two performance metrics to construct plots for visual analysis of performance. These are normalised regret (Section \ref{section:normalised_regret}) and ranking plots (Section \ref{section:metrics_ranking}). We also provide some analysis of the historic datasets for each benchmark (see Section \ref{section:problem_statement} for description of historic datasets), with a view to gaining insight into relative locations of minima in the input domain for different tasks. For this we require a distance metric to compare distance between input data points (Section \ref{section:metrics_hist_data}).\\

\subsubsection{Normalised Regret}
\label{section:normalised_regret}
The first evaluation metric used to compare different methods of BO using ensemble-based transfer learning is normalised simple regret\footnote{This is equivalent to average distance to the global minimum (ADTM) used on grid benchmarks \cite{Wistuba2016two, Wistuba2018scalable}} \cite{Wistuba2016two, Wistuba2018scalable, Pineda2021hpob}. In a transfer learning setting, where we compare function surfaces from related but not identical tasks, different task objective functions may have a different output range over the selected input domain. Therefore, to compare average performance over all tasks, it is necessary to use normalised simple regret. Other examples of work using this metric include \cite{Wistuba2021fewshot, Feurer2022, Wistuba2016two, Wistuba2018scalable, Jomaa2021transfer}. Plots in Section \ref{section:main_results} show mean normalised simple regret over all seed and task combinations for a particular benchmark\footnote{For benchmarks with 38 tasks, and 15 seeds, this produces 570 seed-tasks over which to average normalised regret.}.\\

\subsubsection{Ranking Plots}
\label{section:metrics_ranking}

The second evaluation metric plots we use to compare different ensemble-based transfer learning with BO methods are ranking plots. They were also used in multiple other works including  \cite{Schilling2016, Wistuba2016two, Feurer2018, Wistuba2018scalable, Feurer2022,Pineda2021hpob}\footnote{In this work we prepare these plots by computing rank separately for each seed and task combination and evaluation budget iteration across a set of methods we are comparing. Then for each method we plot the average rank across all seeds and tasks for each iteration.}. The resulting plots can be seen in Section \ref{section:main_results}. These ranking plots provide a different perspective on comparative performance between methods as compared to the normalised regret plots. While normalised regret may be impacted by variable performance across different seeds or tasks for one method compared to another\footnote{If one method exhibits a large amount of variation in performance between different seeds or tasks it will impact the value of average normalised simple regret.}, ranking plots just show the average rank of a particular method over all tasks and seeds, irrespective of how big the difference is (comparing different methods) in performance on a particular seed and task.\\

\subsubsection{Metrics for Historic Dataset Analysis}
\label{section:metrics_hist_data}

In addition to using metrics to compare performance of different BO with ensemble-based transfer learning methods, we also analyse historic datasets across all tasks with the aim of gaining insight into how comparative location of minima in different tasks impacts the performance of transfer learning with BO. Since we include a range of benchmarks with varying dimension and datatypes, comparison of input locations requires a distance metric that can handle different number of dimensions and datatypes. The distance metric we choose is Gower distance \cite{Gower1971general}. To compute Gower distance, a similarity score between $0$ and $1$ is first computed between two points by averaging the similarity between each dimension of the two points over all dimensions (see \cite{Gower1971general} for details). This enables flexibility for different variable types, with exact approach to computing similarity varying for qualitative versus quantitative variables. The simiarlity score is then converted to a distance proportional to $(1-S_{i,j})^{1/2}$, where $S_{i,j}$ is similarity between points $i$ and $j$, as described in \cite{Gower1971general}.\\

Using Gower distance, we can also obtain a clustering of points using agglomerative clustering or spectral clustering  \cite{Scikit-learn}. For this work we use agglomerative clustering, with distance threshold set to $0.02$ and complete linkage. We also use spectral clustering with the same number of clusters discovered during agglomerative clustering to validate clusters discovered with agglomerative clustering \cite{Xiong2018clustering}. Clustering is experimental and requires validation. There are a number of approaches used, such as verification with other labels \cite{Xiong2018clustering}. However, in this context, task labels and location clusters are independent. This is what we are trying to seek information about. Therefore, we use an alternative clustering algorithm as validation. This approach was chosen as a way to approximate overlapping input points that are minima in the historic dataset. It is based on the idea that if clusters are very small, the input points in them are approximately overlapping. If there exists an overlap of minima location between two tasks' historic datasets (obtained using standard BO), one being a source task and one being the target task, then we assume it is likely that using a surrogate model trained on the source task's historic dataset in the ensemble will lead to the discovery of that minima location in the target task during ensemble-based transfer learning BO. That means that in theory, transfer learning BO should perform at least as well on the target task as standard BO. Note that this is an analytical tool only, and not one that makes sense on a new problem, since for a real problem, you would not have access to the target task's expensive black-box objective function historic dataset, but only to source task historic datasets. \\

Using the clusters obtained as an approximation for data points overlapping, we also propose an approach to computing probability of approximate overlap between minima data points from different tasks' historic datasets. Equation \ref{equation:probability_overlap} shows how we calculate the probability of at least one overlapping minima between the historic dataset for task $i$ and the union of historic datasets from all other tasks. The probability shown here is the probability of overlap for task $i$ out of the $N$ tasks, averaged over all seeds, where $S$ is number of seeds used in the set of historic datasets and $s_{task_i}$ refers to a seeds for task $i$, and $cids_{(task_i,s_{task_i})}$ is the set of cluster ids in which task $i$, seed $s_{task_i}$ has at least one minima datapoint (and likewise for task $j$). An indicator function, $\mathbb{I}$, is $1$ if intersection is the empty set between cluster ids in the task $i$, seed $s_{task_i}$, and task $j$, seed $s_{task_i}$, and other wise $0$\footnote{We are taking the sum over all seeds for task $j$ for which the intersect of cluster sets is empty.}. The product term computes probability of no overlap between task $i$ and other tasks. The probability of at least one overlapping cluster is $1$ minus this amount. The average over all seeds for task $i$ is

\begin{equation}
    \label{equation:probability_overlap}
    p_{overlap_{task_i}} = \frac{1}{S} \sum_{s_{task_i}=1}^S \Bigg[ 1 - \prod_{j=1,j \neq i}^N \Bigg(\frac{\sum_{s_{task_j}=1}^S\mathbb{I}\big(\{cids_{(task_i,s_{task_i})}\} \cap \{cids_{(task_j,s_{task_j})}\} \in \emptyset \big)}{S} \Bigg) \Bigg].
\end{equation}


\section{Main Results}
\label{section:main_results}

This section responds to the three questions posed in Section \ref{section:intro_questions}. We address the first question in section \ref{section:address_q1}. First we analyse the effectiveness of warm start initialisation as compared to random initialisation (Latin hypercube sampling) using normalised regret plots in Section \ref{section:main_results_warm_start}. Then we compare performance of different weighting strategies with both warm start and random initialisation, again using normalised regret plots in Section \ref{section:main_results_weighting}. Then we address the second question by comparing weighting strategy methods with and without handling of poor transfer learning in Section \ref{section:main_results_bad_transfer_learning}. Normalised regret plots addressing this question can be found in Section \ref{section:main_results_bad_transfer_learning_norm_regr}, and in Section \ref{section:ranking_plots} we include ranking plots to provide further analysis of comparative performance of various methods included in this study. In Section \ref{section:results_box_plot} we include a box-plot summary of our analysis of historical data in response to the third question.\\

\subsection{Ensemble-Based Transfer Learning Strategies with Bayesian Optimisation}
\label{section:address_q1}
In this section we present results to address the first question in section \ref{section:intro_questions}. We examine the effectiveness of the warm start initialisation and compare different weighting strategies.

\subsubsection{Effectiveness of Warm Start}
\label{section:main_results_warm_start}

Firstly we compare the effect of using $2$ initialisation points obtained via the warm start algorithm (see Algorithm \ref{algorithm:warm_start}) against the effect of using $10$ random initialisation points obtained using Latin hypercube sampling. The number of initialisation points chosen for each method is selected to be consistent with the implementation in \cite{Feurer2022}. We make this comparison for standard BO, and the ensemble-based transfer learning weighting strategy methods described in Section \ref{section:weights} which includes the following: LaGPE with positive weights constraint, LaGPE without positive weights constraint, RiGPE with positive weights constraint, RiGPE without positive weights constraint, RGPE, TSTR and WAC.\\

\begin{figure}
\includegraphics[width=1.0\textwidth]{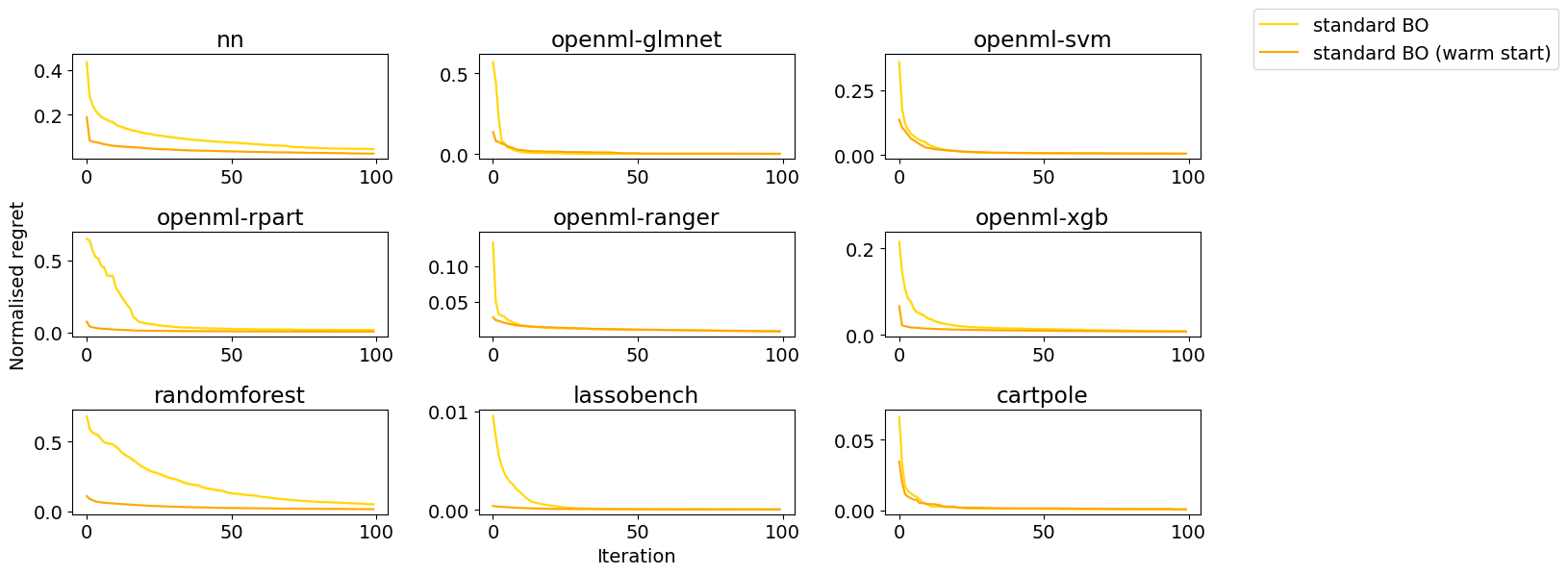}
  \caption{Standard BO plots with different initialisation procedures}
  \label{fig:standard_bo}
\end{figure}

For standard BO (see Figure \ref{fig:standard_bo}), it can be observed that for most benchmarks, performance of BO is improved by using only $2$ warm start initialisation points as compared to $10$ random initialisation points. This is particularly evident during early iterations, where the warm start initialisation points minimse normalised regret values considerably. Examples include for nn, openml-rpart, openml-xgb, randomforest and lassobench benchmarks, where the orange line representing warm start initialisation is much lower than the yellow line representing random initialisation. While for benchmarks openml-svm, openml-ranger and cartpole the performance of standard BO with warm start initialisation is not significantly better than with random initialisation, over the 100 iterations, it is comparable. The only benchmark where random initialisation with standard BO significantly out-performed the version with warm start initialisaiton was openml-glmnet.\\

\begin{figure}
\includegraphics[width=1.0\textwidth]{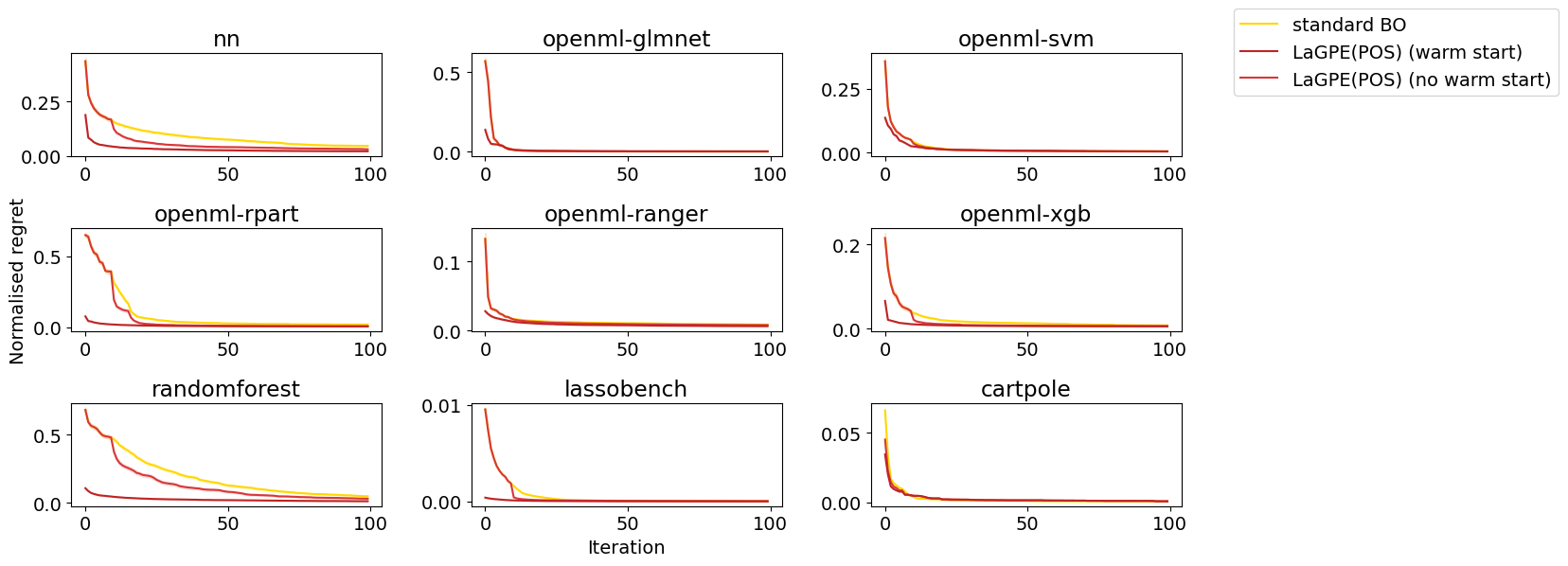}
    \caption{LaGPE plots with different initialisation procedures}
    \label{fig:lagpe_bo}
\end{figure}

Figure \ref{fig:lagpe_bo} shows results for the LaGPE weighting strategy with weights constrained to be positive. It can be seen that 6 out of 9 benchmarks (nn, openml-rpart, openml-ranger, openml-xgb, randomforest and lassobench) show improved BO performance over all 100 iterations when warm start initialisation is used in combination with the LaGPE weighting strategy as compared to random initialisation with the LaGPE weighting strategy or standard BO. For openml-glmnet, LaGPE with random initialisation is best during early iterations, but the warm start initialisation with LaGPE version works better in later iterations, and performs almost as well as standard BO over the 100 iterations. For openml-svm, all 3 methods show similar performance over the 100 iterations, and for cartpole, standard BO was better for most iterations than both LaGPE method versions (with warm start initialisation and with random initialisation), although by the 100th iteration, all 3 are very similar.\\

\begin{figure}
\includegraphics[width=1.0\textwidth]{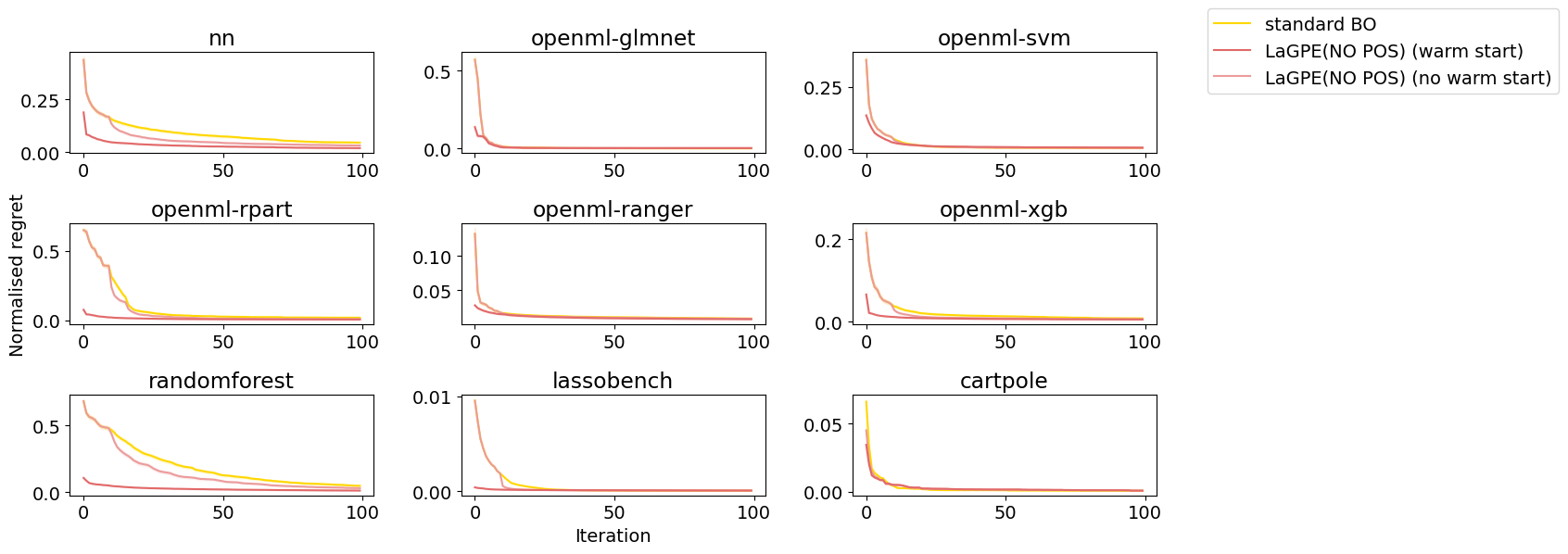}
    \caption{LaGPE (no constraint on weights) plots with different initialisation procedures}\label{fig:lagpe_np_bo}
\end{figure}

Similar results can be seen for the LaGPE (without positive weights constraint) weighting strategy in Figure \ref{fig:lagpe_np_bo}. For this method, 5 out of 9 benchmarks, nn, openml-rpart, openml-ranger, openml-xgb and randomforest, show significantly improved optimisation performance using warm start initialisation with LaGPE (without positive weights constraint) as compared to random initialisation with LaGPE (without positive weights constraint) or standard BO. Out of the remaining benchmarks, openml-glmnet, openml-svm and lassobench show best performance for LaGPE (without positive weights constraint) with warm start during initial iterations, but best performance for standard BO during later iterations. For cartpole, standard BO was the best performing method for all iterations.\\

\begin{figure}
\includegraphics[width=1.0\textwidth]{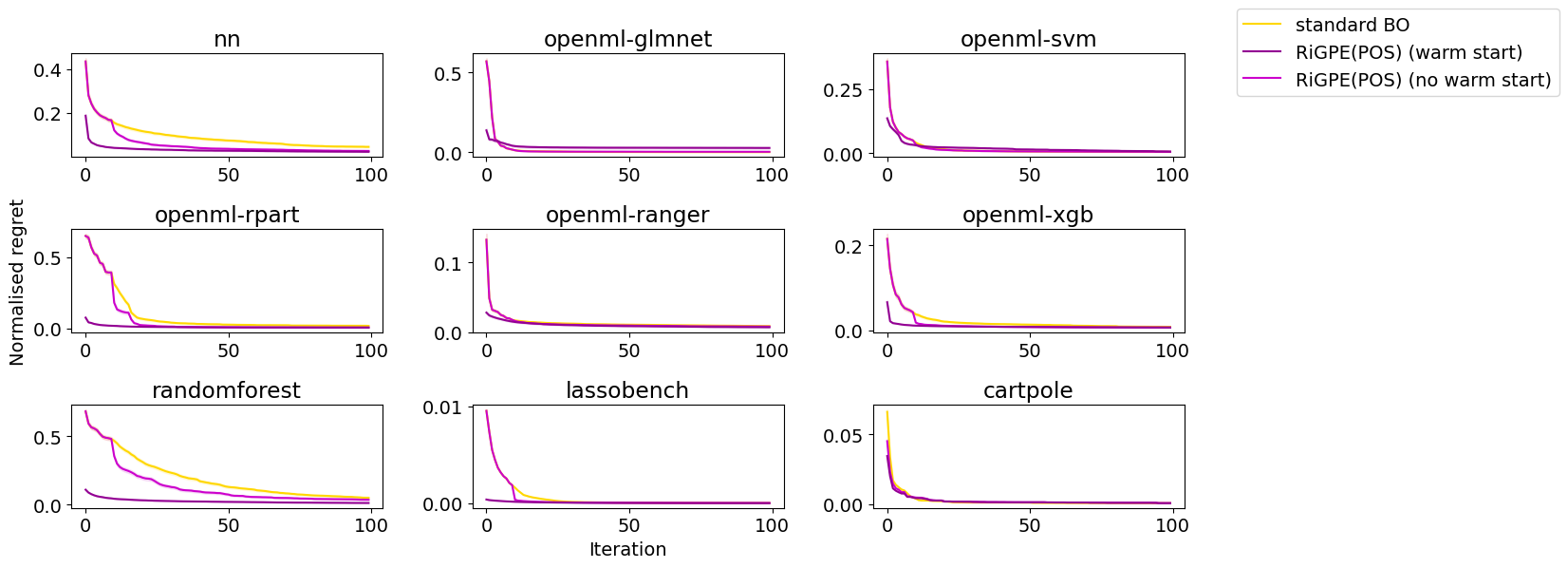}
    \caption{RiGPE (positive weights constraint) plots with different initialisation procedures}
    \label{fig:rigpe_bo}
\end{figure}

\begin{figure}
\includegraphics[width=1.0\textwidth]{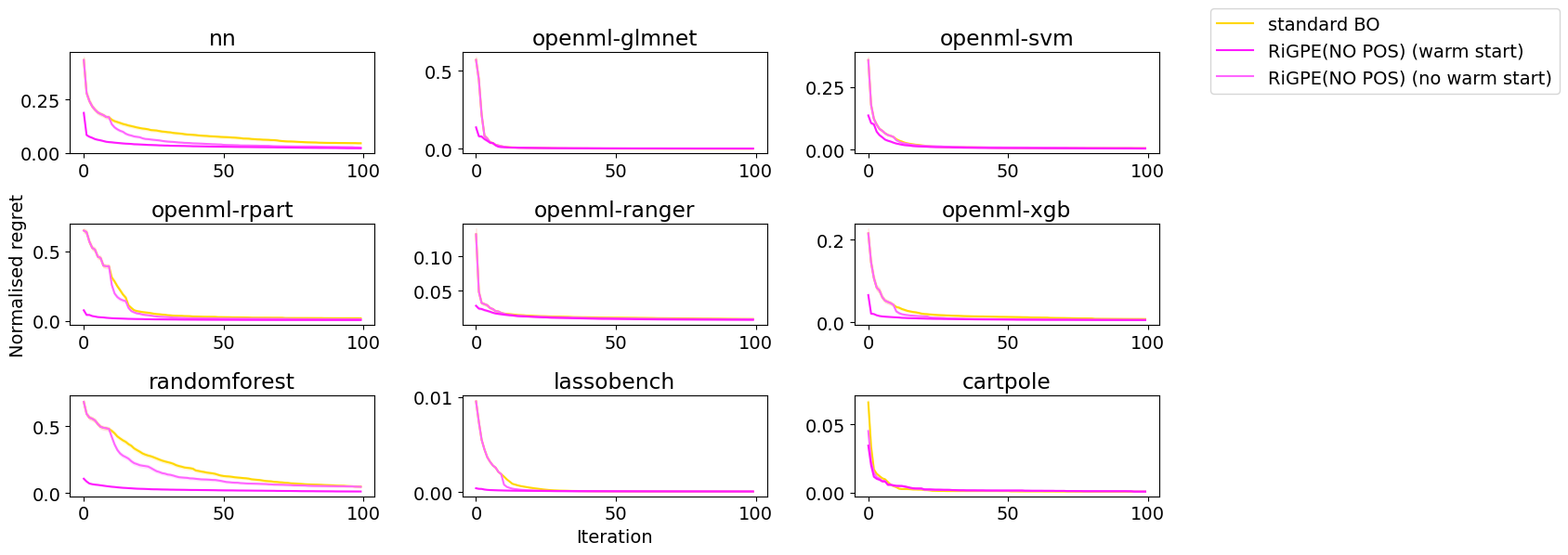}
    \caption{RiGPE (no weights constraint) plots with different initialisation procedures}
    \label{fig:rigpe_np_bo}
\end{figure}

Then we look at the Ridge regression weighting strategy methods in Figure \ref{fig:rigpe_bo} with positive weights constraint and Figure \ref{fig:rigpe_np_bo} without positive weights constraint. For Ridge regression weighting strategy with positive weights constraint, 6 out of 9 benchmarks, nn, openml-rpart, openml-ranger, openml-xgb, randomforest, and lassobench, the warm start initialisation method show better overall optimisation performance than the random initialisation method, or standard BO. For benchmarks openml-glmnet and openml-svm, method without warm start initialisation (RiGPE and standard BO) show better optimisation performance. For cartpole, standard BO was again the best method of optimisation.\\

For Ridge regression weighting strategy without positive weights constraint, again 6 out of 9 benchmarks, nn, openml-svm, openml-rpart, openml-ranger, openml-xgb, and randomforest showed better performance when warm start initialisation was used as compared to random initialisation or standard BO. This time, for lassobench, the warm start initialisation method performed best over the first few iterations, but then the random initialisation and standard BO methods improved more quickly. For openml-glmnet, standard BO was slightly better over most iterations (random initialisation was better initially), and for cartpole, standard BO was better than the RiGPE methods until the very last few iterations.\\

\begin{figure}
\includegraphics[width=1.0\textwidth]{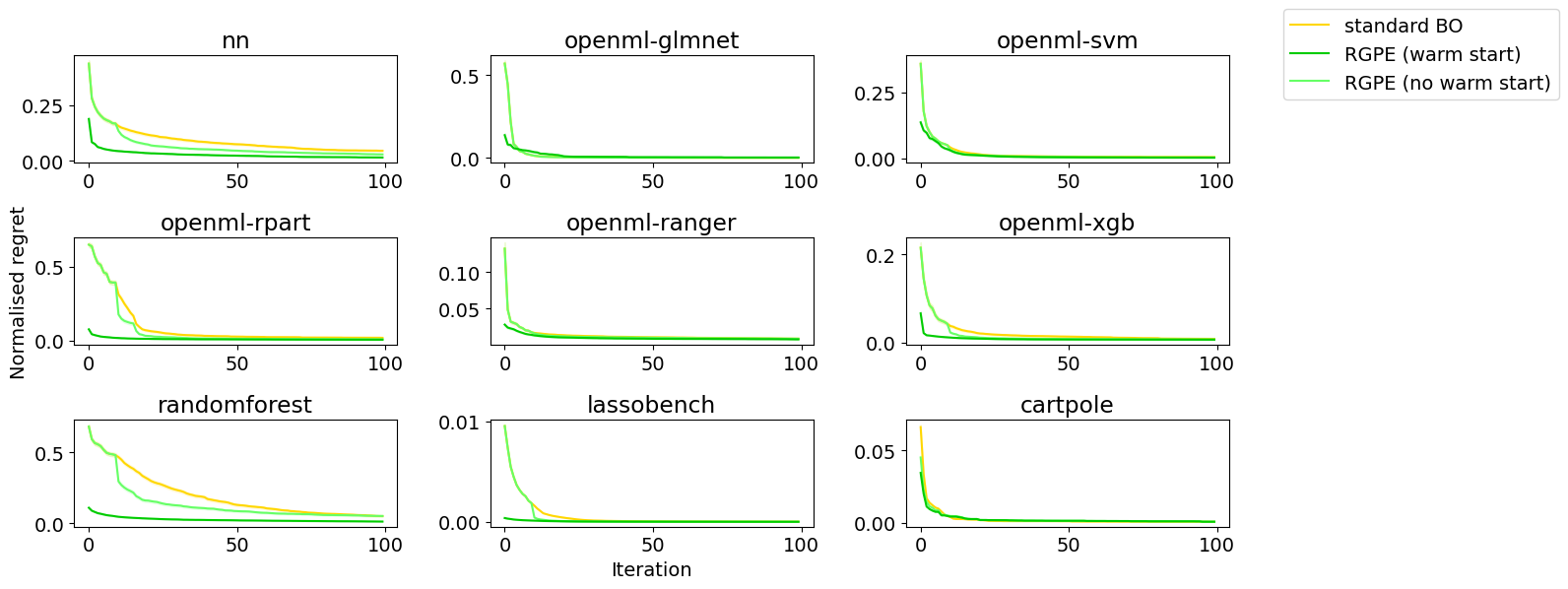}
    \caption{RGPE plots with different initialisation procedures}
    \label{fig:rgpe_bo}
\end{figure}

Then we compare the effect of the two initialisation approaches for the RGPE weighting scheme in Figure \ref{fig:rgpe_bo}. This time, 7 out of the 9 benchmarks show best performance over the 100 iterations with the warm start initialisation. For openml-glmnet, BO performance was better without the warm start initialisation, and cartpole BO performance was similar for both.\\

\begin{figure}
\includegraphics[width=1.0\textwidth]{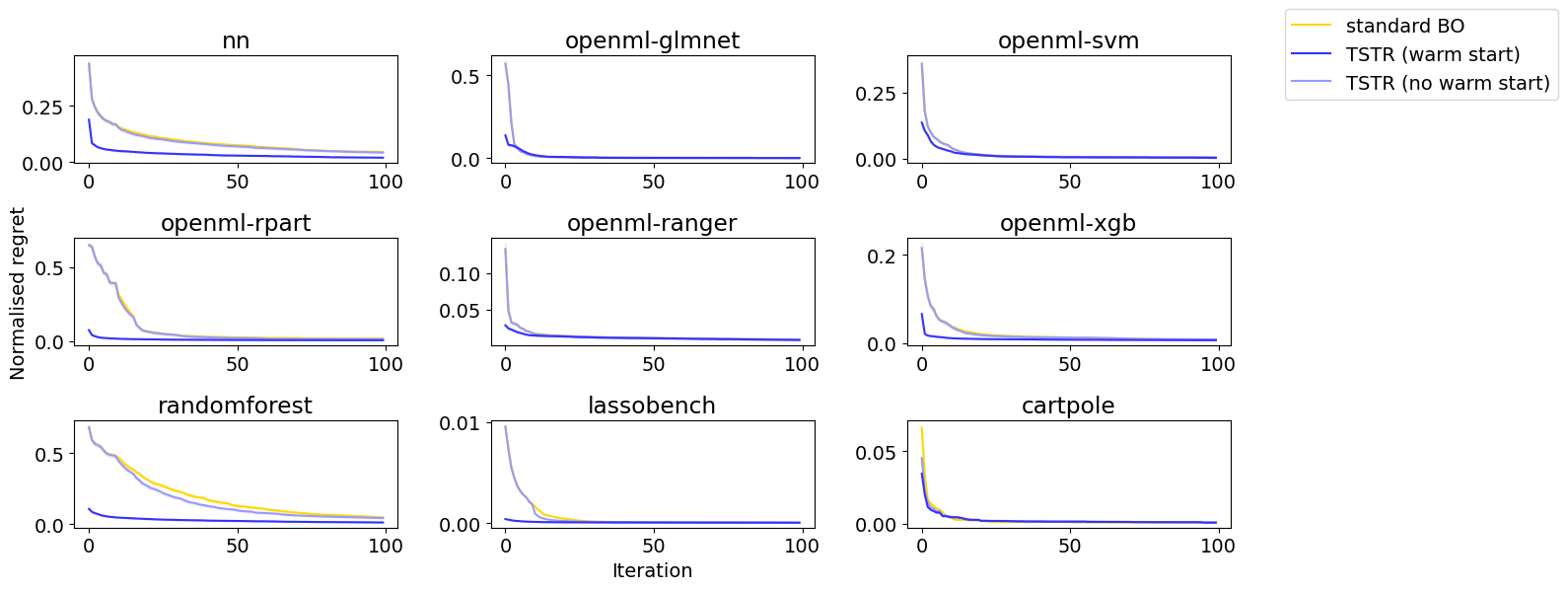}
    \caption{TSTR plots with different initialisation procedures}
    \label{fig:TSTR_bo}
\end{figure}

The TSTR method for weighting can be seen in Figure \ref{fig:TSTR_bo}. For 8 out of 9 benchmarks, these plots show that the warm start initialisation performs at least as well as other methods. In particular during early iterations, using warm start initialisation significantly improves performance of BO as compared to using random initialisation.\\

\begin{figure}
\includegraphics[width=1.0\textwidth]{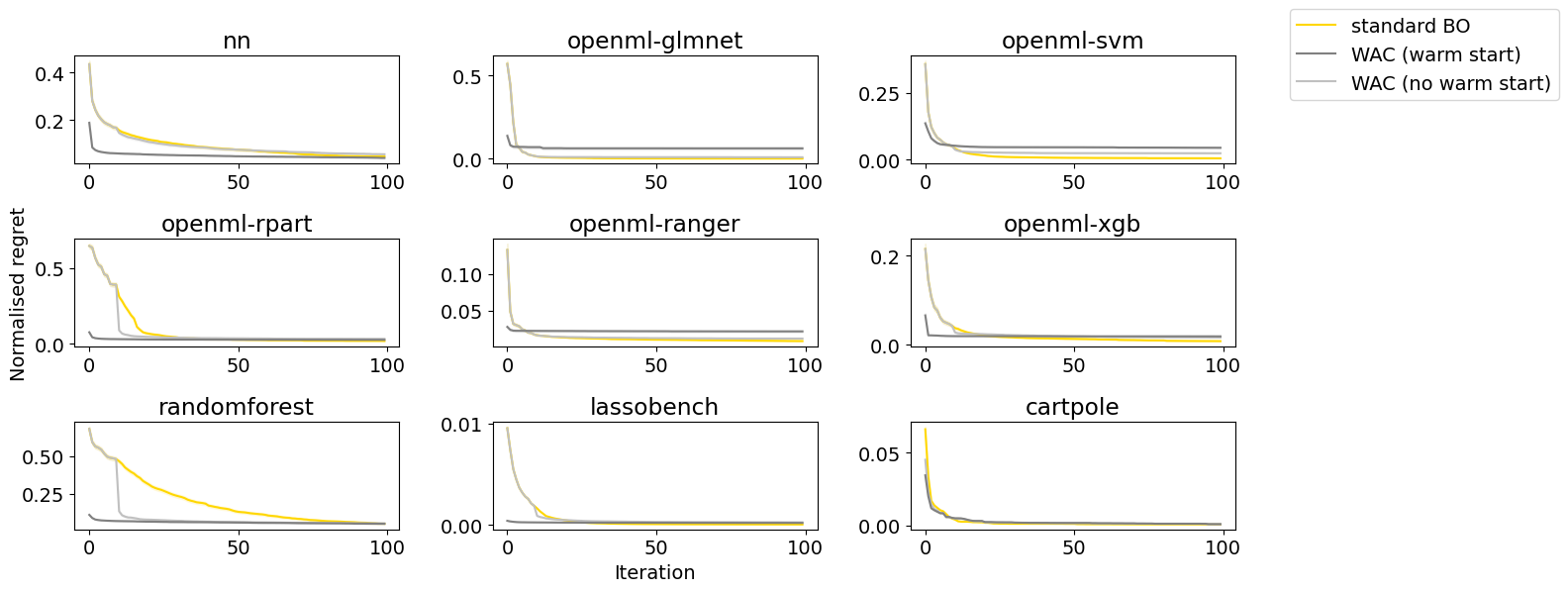}
    \caption{WAC plots with different initialisation procedures}\label{fig:wac_bo}
\end{figure}

Finally we look at the WAC weighting strategy in Figure \ref{fig:wac_bo}. Again during early iterations the warm start initilisation improves performance of BO. However, this improvement only persists throughout the 100 iterations for 2 out of 9 benchmarks, nn and randomforest. For other benchamrks, standard BO performs better than both WAC with warm start initialisation and random initialisation, suggesting that in many cases standard BO may out-perform the WAC weighting strategy method of BO.\\

\subsubsection{Comparing Different Weighting Strategies}
\label{section:main_results_weighting}

\begin{figure}
\includegraphics[width=1.0\textwidth]{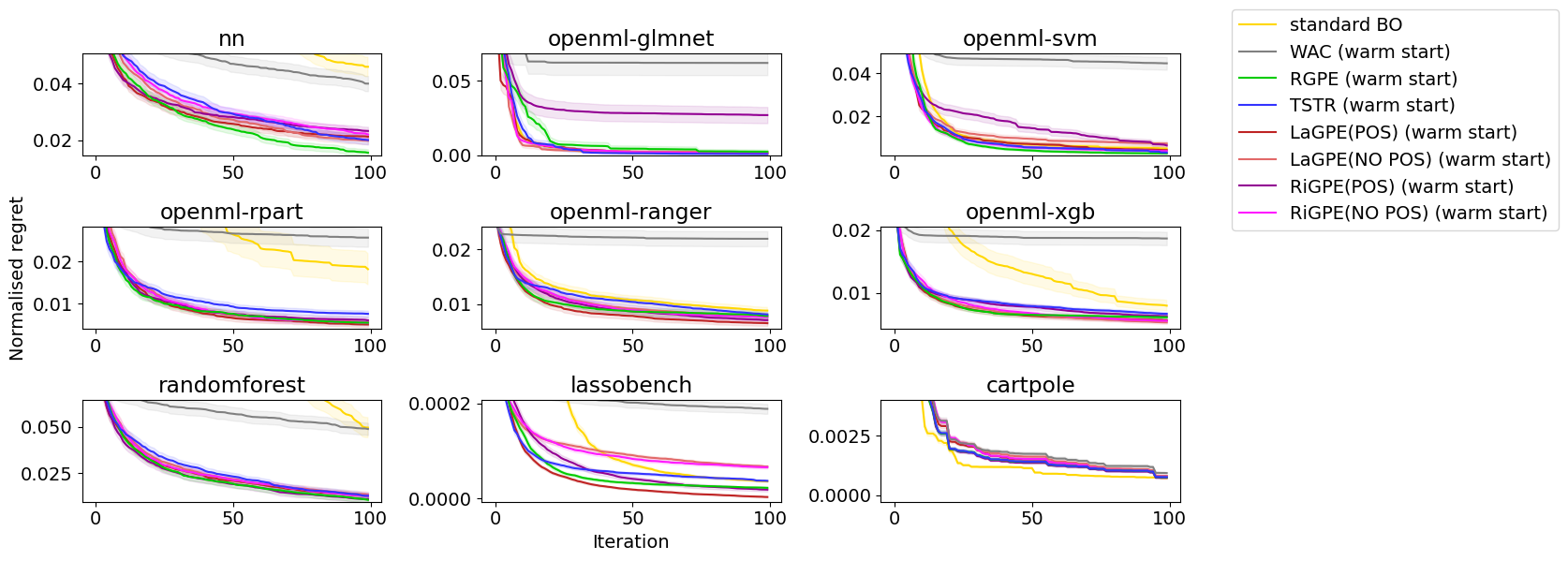}
    \caption{Plots comparing different weighting strategies using warm start initialisation}\label{fig:ws_weighting_plots}
\end{figure}

In this section we look at how the different weighting strategies described in the the previous section (\ref{section:main_results_warm_start}) compare with each other and standard BO. We include two versions of our weighting strategies. In the first version in Figure \ref{fig:ws_weighting_plots}, all weighting strategies use $2$ warm start initialisation points. (Please note, these plots have been zoomed in more than the plots in Section \ref{section:main_results_warm_start}.) For 2 out of 9 benchmarks, nn and openml-svm, BO with RGPE (without weight dilution prevention strategy) shows best performance. For another 2 out of the 9 benchmarks, openml-ranger and lassobench, BO using LaGPE (POS) (Lasso with positive weights constraint without alternating strategy) shows best performance, and for 1 out of the 9 benchmarks, cartpole, standard BO was the best optimisation strategy. For the other 4 out of 9 benchmarks, openml-glmnet, openml-rpart, openml-xgb and randomforest, various weighting strategies performed similarly and it is difficult to select the best. Also which weighting strategies were best on these 4 benchmarks varied throughout the iterations, with no clear winner overall. Given results overall, methods such as RGPE and LaGPE (POS) that enforce positive weights seem to be better than other methods.\\

\begin{figure}
\includegraphics[width=1.0\textwidth]{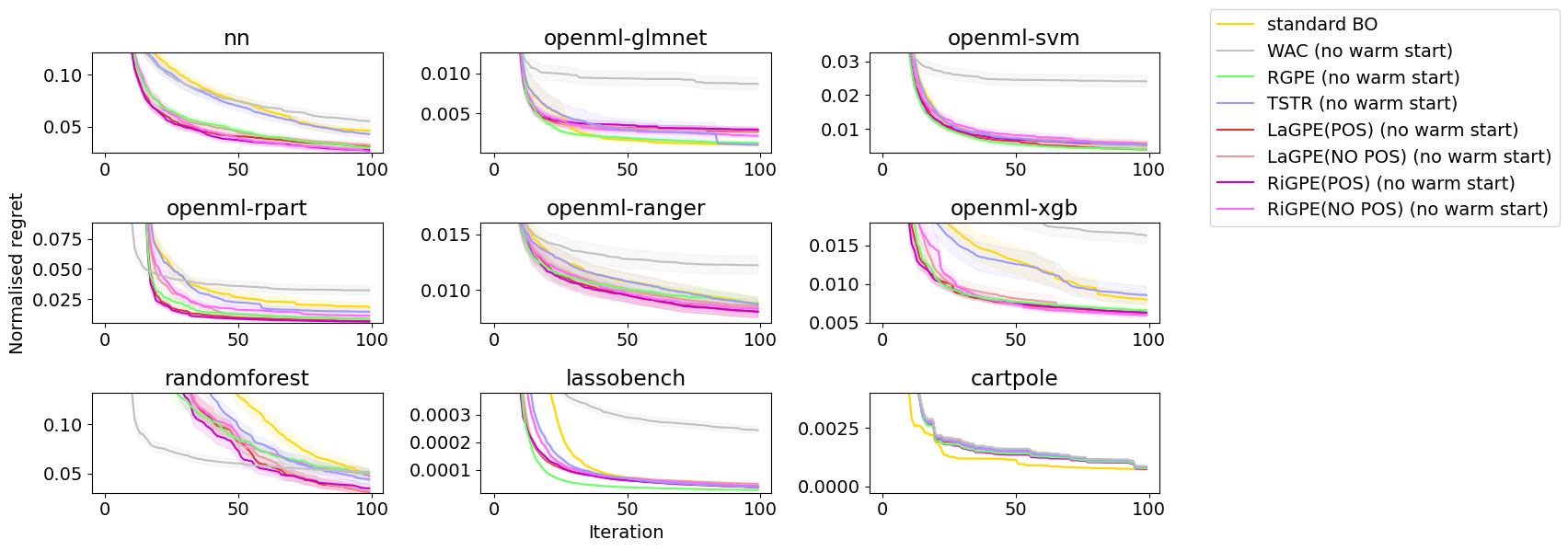}
    \caption{Plots comparing different weighting strategies using random initialisation}\label{fig:no_ws_weighting_plots}
\end{figure}

For plots showing comparison of the different weighting strategies when random initialisation ($10$ points) is used (in Figure \ref{fig:no_ws_weighting_plots}), the majority of methods perform similarly overall across most benchmarks. While for lassobench RGPE is better than other methods overall, and for cartpole standard BO is better than other methods overall, for other benchmarks is it difficult to see one method that outperforms the others over the 100 iterations. Since we have already shown in the previous section (\ref{section:main_results_warm_start}) that, in general, BO methods using the warm start initialisation perform better, there is no need to further analyse plots showing the random initialisation versions.\\

\subsection{Comparing Methods With and Without a Strategy to Handle Bad Transfer Learning}
\label{section:main_results_bad_transfer_learning}

In this section we address the second question in section \ref{section:intro_questions} by exploring how the inclusion of a strategy for automated selection of standard BO versus transfer learning BO, based on measurable criteria available during optimisation, affects the overall performance of a Bayesian optimisation pipeline. The exact strategy used depends on the weighting strategy component of the pipeline as described in Section \ref{section:methods_handling_worst_case}.\\

\subsubsection{Comparing Methods With and Without a Strategy to Handle Bad Transfer Learning Using Normalised Regret Plots}
\label{section:main_results_bad_transfer_learning_norm_regr}

\begin{figure}
\includegraphics[width=1.0\textwidth]{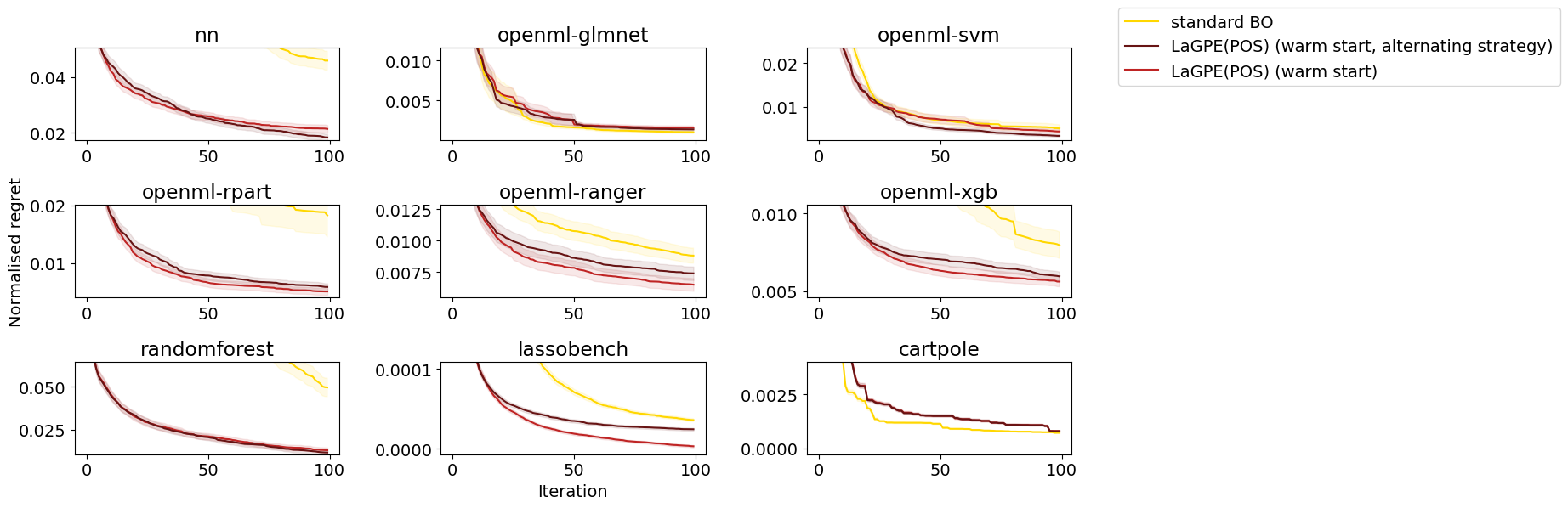}
    \caption{LaGPE (with positive constraint on the weights) plots with and without strategy to handle bad transfer learning}\label{fig:lagpe_bad_tl}
\end{figure}

For comparing performance with and without the component to handle bad transfer learning, we have included for analysis only the plots for weighting strategies that showed best performance for one or more benchmarks (see Figure \ref{fig:full_plots} in Appendix \ref{appendix:norm_regret_bad_tl}). The relevant plots are found in Figure \ref{fig:lagpe_bad_tl} for LaGPE with positive constraint, Figure \ref{fig:lagpe_np_bad_tl} for LaGPE without positive constraint and Figure \ref{fig:rgpe_bad_tl} for RGPE. (See Appendix section \ref{appendix:norm_regret_bad_tl} for plots with RiGPE with positive constraint, RiGPE without positive constraint, and TSTR weighting strategies.) We do this to understand how this component affects performance empirically on the benchmarks included in this study.\\

It can be seen in Figure \ref{fig:lagpe_bad_tl} that the addition of our proposed component to handle bad transfer learning with the LaGPE weighting strategy has little impact on overall performance. For 4 out of 9 benchmarks, openml-rpart, openml-ranger, lassobench and openml-xgb, methods without the alternating strategy perform better overall. For 3 out of the 9 benchmarks, openml-glmnet, randomforest and cartpole, the alternating strategy makes little difference to the performance, and for the remaining 2 benchmarks, nn and openml-svm, the alternating strategy provides small improvements, as compared to methods without, in later iterations only.\\

\begin{figure}
\includegraphics[width=1.0\textwidth]{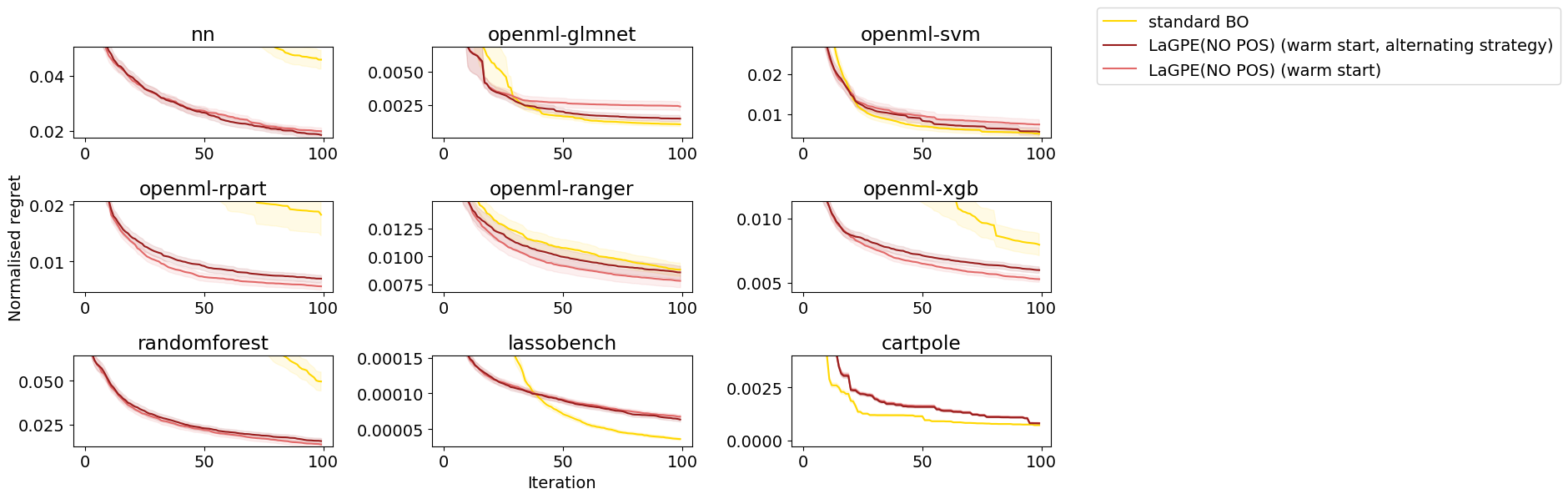}
    \caption{LaGPE (without positive constraint on the weights) plots with and without strategy to handle bad transfer learning}\label{fig:lagpe_np_bad_tl}
\end{figure}

In Figure \ref{fig:lagpe_np_bad_tl}, showing plots for LaGPE (without positive weights constraint) methods, results are similar as for LaGPE (with positive weights constraint). For 3 out of 9 benchmarks, openml-rpart, openml-ranger, and openml-xgb, methods without the alternating strategy clearly performed better. For 4 out of 9 benchmarks, nn, randomforest, lassobench and cartpole, the alternating strategy made no real difference, and for the remaining 2 out of 9 benchmarks, openml-glmnet and openlm-svm, there was some improvement in methods using the alternating strategy.\\

\begin{figure}
\includegraphics[width=1.0\textwidth]{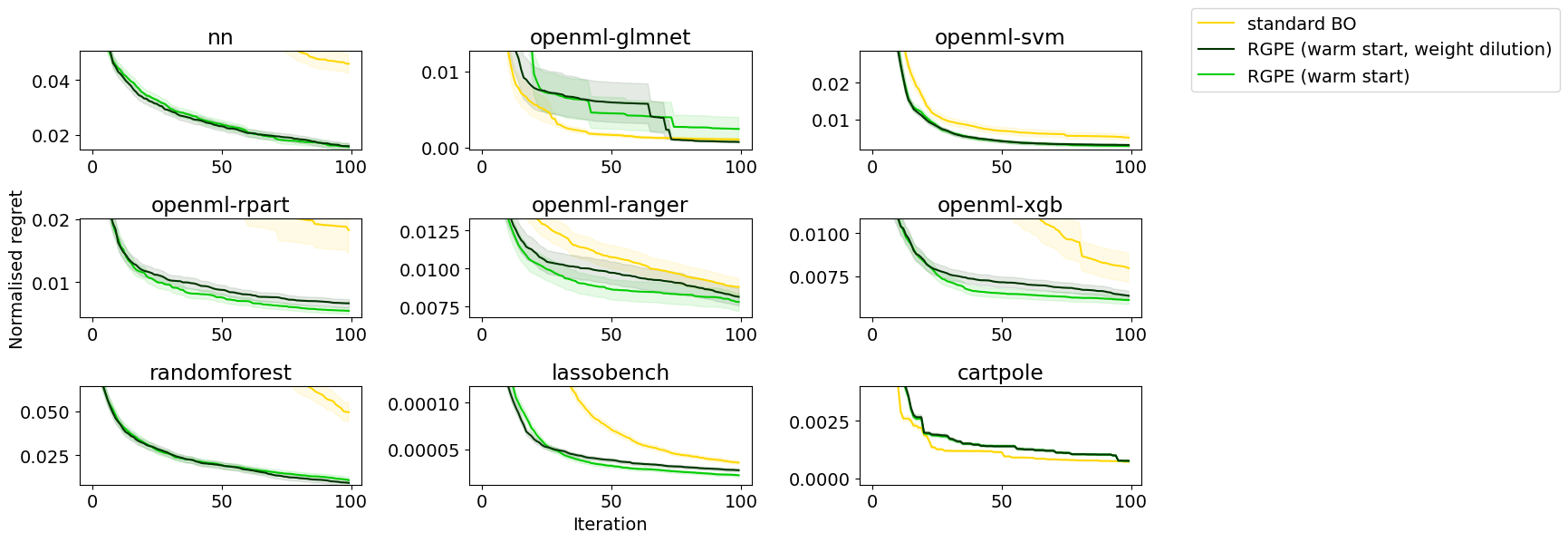}
    \caption{RGPE plots with and without strategy to handle bad transfer learning}\label{fig:rgpe_bad_tl}
\end{figure}

Again, for methods using the RGPE weighting strategy in Figure \ref{fig:rgpe_bad_tl}, the component in the pipeline to handle bad transfer learning does not significantly improve overall performance. In 4 out of 9 benchmarks, openml-rpart, openml-ranger, lassobench and openml-xgb, methods without weight dilution prevention perform better over the 100 iterations. For 4 out of 9 benchmarks, nn, openml-svm, randomforest and cartpole, the weight dilution prevention strategy appears to make no difference. Only for 1 out of the 9 benchmarks, openml-glmnet, does the weight dilution prevention strategy improve perfomance, and only in later iterations. It is interesting to note that the weight dilution prevention strategy to selectively eliminate source surrogate models (described in Section \ref{section:methods_handling_worst_case}), did not appear to perform significantly better than our more abrupt proposed method for handling bad transfer learning. While there are definitely regions of optimisation where standard BO will perform better, as can be seen during early iterations of openml-glmnet in Figure \ref{fig:rgpe_bad_tl}, it is challenging to design an automated approach to detect and exploit this during optimisation.\\

\subsubsection{Comparing Methods With and Without a Strategy to Handle Bad Transfer Learning Using Ranking Plots}
\label{section:ranking_plots}

\begin{figure}
\includegraphics[width=1.0\textwidth]{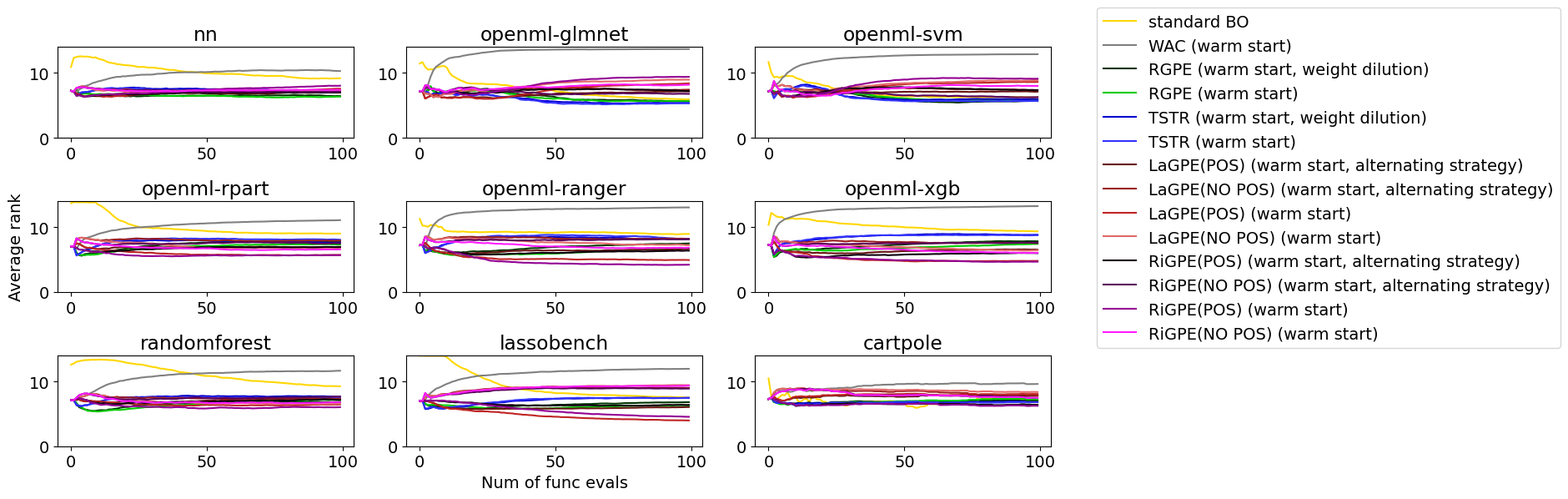}
    \caption{Plots comparing different methods' average rank over evaluation budget (lower rank indicates better performing method)}
    \label{fig:ranking}
\end{figure}

In analysing the impact of the inclusion of a strategy for handling bad transfer learning in the transfer learning BO pipeline we also examine the ranking plots in Figure \ref{fig:ranking}. These plots give an indication of how different methods compare in relation to each other, on average, without being affected by the size of the difference in normalised regret\footnote{For example, if there is one particular task in a benchmark for which a particular method performs much worse on as compared to other tasks, it will affect the average normalised regret (which is affected by the size of difference) more than the averaged rank (which is affected by the rank integer only)}. It can be seen that in general, results are similar to normalised regret plots, with some exceptions.\\ 

For 1 out of 9 benchmarks, nn, RGPE (with or without weight dilution prevention) is the best ranking method overall. For 1 out of 9 benchmarks, openml-svm, RGPE without weight dilution, and TSTR with or without weight dilution prevention are the best ranking methods overall. For 1 out of 9 benchmarks, openml-glmnet, TSTR, with or without weight dilution prevention, is the best ranking method. For 2 out of 9 benchmarks, openml-rpart and openml-xgb, LaGPE(POS) and RiGPE(POS), both without the alternating strategy, are the best ranking methods overall. For 2 out of 9 benchmarks, openml-ranger and randomforest, RiGPE(POS) without the alternating strategy is best ranking method. For 1 out of 9 benchmarks, lassobench, LaGPE(POS) without alternating strategy is the best ranking method. For cartpole, best ranking method varies between standard BO and RiGPE(POS) without the alternating strategy, depending on iteration number.\\ 

For all benchmarks it can be seen that during early iterations there is a lot of variation in how different methods rank, where as during later iterations it is clearer which method performs best. Also, for some benchmarks such as randomforest, nn and cartpole, the majority of methods have similar rank (ie. comparative performance may be more dependent on task number than the method itself), where as for other benchmarks such as openml-ranger and lassobench there is greater variation in method rankings suggesting that for these benchmarks, particular methods do perform differently overall.\\

General trends include better performance for methods that do not include strategies to handle bad transfer learning, the WAC method consistently ranks worse than other methods, and regularised regression methods (LaGPE and RiGPE) perform better when weights are constrained to be positive.\\

In general, openml-rpart, openml-ranger, openml-xgb and lassobench perform well with methods that use a regularised regression weighting strategy, where as nn, openml-glmnet and openml-svm work well with a ranking based weights strategy. For randomforest results are not consistent for metrics, and while normalised regret models show better performance for RGPE methods, the ranking plot shows Ridge and Lasso regression models with positive weights constraint to rank better on average, particularly later during the evaluation budget, the ranking based methods. For cartpole, the value of transfer learning in general is not clear, with both normalised regret and ranking plots showing very competitive performance for the standard BO method. Overall, using positive weights seems to help performance, with positive weights constraint versions of regularised regression, as well as RGPE and TSTR where weights are also guaranteed to be positive, being the best performing weighting strategies.\\

\subsection{Analysis of Historic Datasets}
\label{section:results_box_plot}

\begin{figure}
\includegraphics[width=1.0\textwidth]{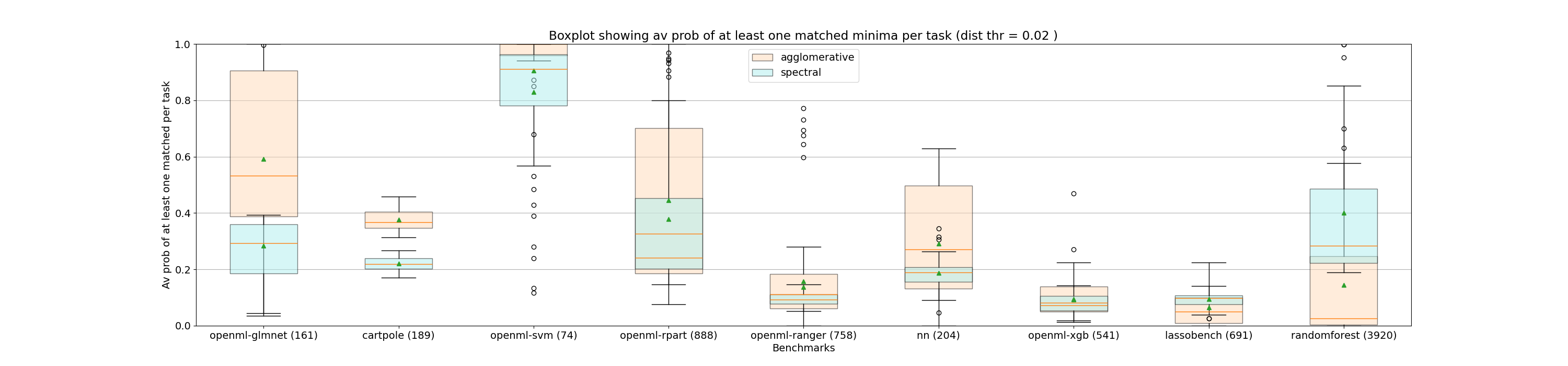}
  \caption{Boxplot showing distribution of probabilities over all tasks across all benchmarks. Benchmarks are ordered along the x-axis with 2 dimensional input on the left, increasing to 10 dimensional input on the right.}
  \label{fig:boxplot}
\end{figure}

The third question in Section \ref{section:intro_questions} is about trying to predict whether ensemble-based transfer learning methods will improve performance of BO as compared to using standard BO, and is based on only information available in historic datasets. In particular, we seek to gain insight into whether similar locations for minima between source and target functions in a benchmark (or minima discovered in historic datasets) implies better performance using ensemble-based transfer learning methods with BO as compared to when minima locations are not similar between source and target functions. Since, when deciding whether to use standard BO or ensemble-based transfer learning BO methods, we assume access only to historic datasets (see Problem Satement in Section \ref{section:problem_statement}), and not the source functions themselves, we have designed our analysis to use only historic datasets from the benchmarks. We first filter all historic datasets to obtain minima points only, and then iteratively designate each task as target task with remaining tasks as source tasks in order to obtain an averged probability of approximate location overlap of minima between source and target tasks for a particular benchmark using the approach described in Section \ref{section:metrics_hist_data} and Equation \ref{equation:probability_overlap}. In filtering historic datasets to obtain the minima, we observe that there are multiple minima locations in each seed-task historic dataset where evaluations of the black-box objective function gave equal minimum values for different inputs. The box-plots showing this probability across all benchmarks on the x-axis can be seen in Figure \ref{fig:boxplot}. The y-axis is the probability of at least one overlap for a particular task. We chose to use boxplots to visualise the distribution of probabilities over all tasks in each benchmark. We present plots for both clustering algorithms, agglomerative and spectral, used to approximate locations to enable the reader to see how clustering output compares, given that clustering is an experimental method requiring verification (see Section \ref{section:metrics_hist_data} for discussion about this).\\

For benchmarks such as cartpole, openml-ranger and openml-xgb it can be seen that the probability of any particular task historic dataset having at least one overlapping minima with all other historic datasets is roughly similar for all tasks, where as for benchmarks such as openml-glmnet openml-rpart or randomforest, there is more difference between tasks in terms of how well their minima locations relate to other historic dataset minima locations. It can also be seen that, in general, benchmarks, nn, openml-rpart, randomforest, where standard BO performs significantly worse than transfer learning BO methods in both the normalised regret plots (see Figure \ref{fig:ws_weighting_plots}) and the ranking plots (see Figure \ref{fig:ranking}), show at least some tasks with high probability of at least one overlapping minima in the boxplots in Figure \ref{fig:boxplot}, suggesting that information in the historic datasets improves BO performance.\\

However, the trend is not consistent for all benchmarks. For example, openml-svm appears to be an exception, with very high probability of overlapping minima, but similarity in performance between standard BO and transfer learning BO methods in both normalised regret and ranking plots. This may be partly due to effective dimensionality of openml-svm being low (1d or 2d) due to conditionality of the categorical variable (see comments in Section \ref{section:main_results_bad_transfer_learning}). Also, for openml-xgb, performance of standard BO is significantly worse than ensemble-based transfer learning BO methods in both normalised regret plots (see Figure \ref{fig:ws_weighting_plots}) and the ranking plots (see Figure \ref{fig:ranking}), but the boxplot in Figure \ref{fig:boxplot} suggests low probability of overlapping minima between target and source functions, which would be expected to lead to comparatively worse performance of the ensemble-based transfer learning BO methods as compared to standard BO. Note that for two dimensional benchmarks, openml-glmnet and cartpole, standard BO method is also very competitive with transfer learning BO methods, particularly in later iterations of the evaluation budget. In summary, it is evident that our approach to historic data analysis still requires refinement. Use of clustering in this approach is still experimental and consequently may be leading to instability in results (see Section \ref{section:metrics_hist_data} for details). Scatter plots for clusters by tasks can be found in the Appendix (\ref{appendix:scatter_plots_historic_data}) in Figure \ref{fig:cluster_scatter_plots}.\\

\section{Discussion \& Conclusion}
\label{section:conclusion}

\subsection{Summary}

In this study we aim to empirically compare performance of a range of ensemble-based transfer learning BO pipelines with standard BO. Included are comparisons of specific components of the transfer learning BO pipeline. We explore results over nine benchmarks with dimensions ranging from $2$ to $10$, some with mixed variables and others with all continuous variables. We also perform analysis on the historic datasets used in transfer learning to gain insight into the relationship between location of minima in different tasks, and comparative performance of ensemble-based transfer learning BO. We use several different evaluation metrics including normalised simple regret plots, ranking plots, and an approach using Gower distance and clustering to approximate probabilities of overlapping minima between task historic datasets.\\

In analysing results, we observe two key trends in performance of transfer learning BO pipeline components. Firstly, \emph{for initialisation of the optimisation process, warm start initialisation using an algorithm proposed in \cite{Lindauer2018warmstarting} in general improves performance of transfer learning BO as compared to using random initialisation over an evaluation budget of $100$ iterations}. Secondly, it is observed that for most benchmarks, \emph{a weighting strategy that forces weights to be positive is more effective than strategies that allow negative weights}. Methods that only allowed positive weights include LaGPE (positive weights constraint), RiGPE (positive weights constraint), RGPE and TSTR. Finally, there is an observable trend towards lower dimensional benchmarks, and those with a smaller search space (categorical or integer variables with less than 10 values) exhibiting better BO performance with ranking based loss functions for computing best weights, and higher dimensional benchmarks with less categorical or integer variables exhibiting better BO performance with regularised regression approaches to computing best weights. \\

\begin{figure}
\includegraphics[width=1.0\textwidth]{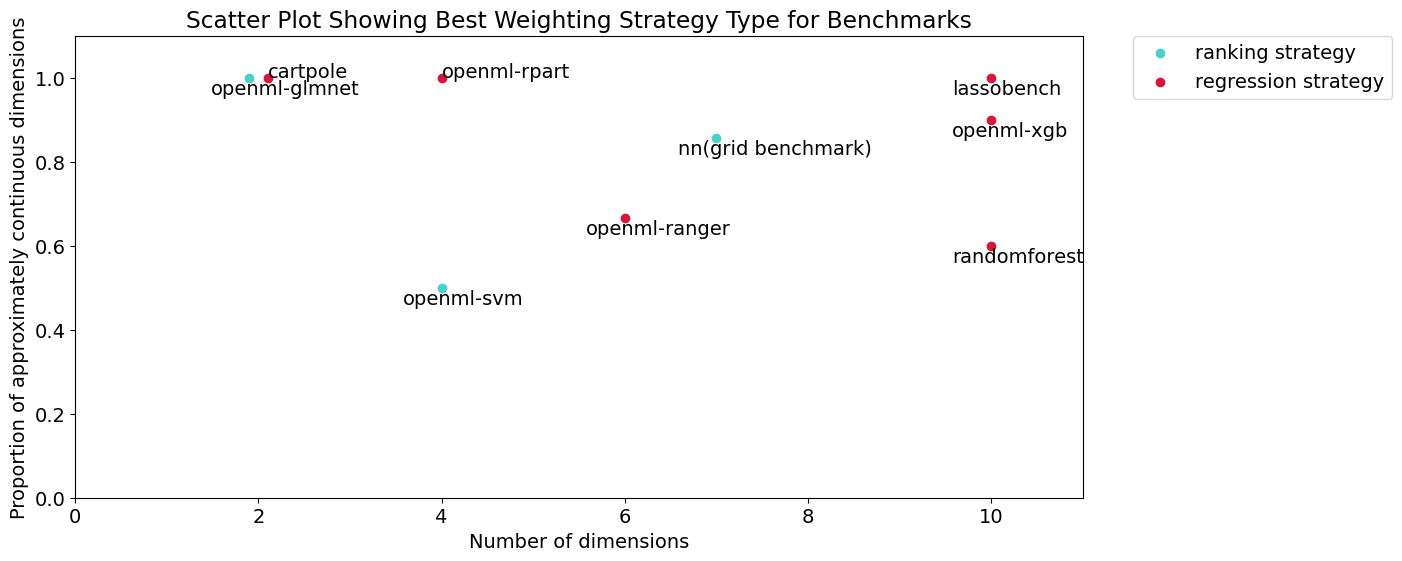}
  \caption{Scatter plot showing relationship between number of dimensions and proportion of approximately continuous variables (> 10 possible values for integer variables) for benchmarks. (Cartpole is excluded since standard BO worked best.)}
  \label{fig:scatterplot_overview}
\end{figure}

The scatter plot in Figure \ref{fig:scatterplot_overview} shows each benchmark in terms of number of dimensions (x-axis), against proportion of approximately continuous dimensions (y-axis). The proportion of approximately continuous dimensions is defined as the proportion of variables that have at least 10 possible values (includes continuous variables). In this plot, benchmark points are coloured by category of weighting strategy that worked best overall in later iterations (approximately 30-100) of the ranking plots in Figure \ref{fig:ranking}. According to the ranking plots, overall best BO performance of benchmarks openml-glmnet, openml-svm, and nn is observed when a ranking based weighting strategy is used to compute best weights with ensemble-based transfer learning, and overall best BO performance on benchmarks cartpole, openml-rpart, openml-ranger, openml-xgb, randomforest and lassobench is observed when regularised regression weighting strategies are used to compute best weights. There is one benchmark, nn, with 7 dimensions, that appears to contradict the trend noted between dimensions and type of weighting strategy. A possible explanation for this is that it is a grid benchmark. While there are 4 continuous variables, the search space is drastically reduced due to limited number of available evaluations (2000)\footnote{This is roughly equivalent to 3 possible values per dimensions for 7 dimensions ($3^7 = 2187$), which is much less than the cut off of 10 possible values for approximately continuous variables in the plot. While we have shown proportion of approximately continuous variables for nn benchmark 0.8, inreality it would be less than that due to it being a grid benchmark.}.\\

\emph{Less clear is the benefit of including a component in the ensemble-based transfer learning with BO pipeline to handle bad transfer learning}. While this component is an important part of performance guarantees provided by \cite{Feurer2022}\footnote{In \cite{Feurer2022}, a guarantee is provided that the ensemble-based transfer learning with BO will not be slower than some multiple of standard BO.}, in our empirical analysis, it did not appear to provide significant improvement to performance. This was consistent with the equivalent component we proposed for use with the regularised regression weighting strategies. Also less clear is exactly \emph{which weighting strategy will work with a particular benchmark}. While there is a vague trend towards ranking based weighting strategies working with lower dimensional benchmarks (openml-glmnet, openml-svm) and/or benchmarks with more categorical variables (randomforest) or discretized input domains (including nn which is a grid benchmark, making it disretized in practice), and regularised regression based weighting strategies working better with benchmarks that have a higher number of dimensions and/or less categorical variables (openml-rpart, openml-ranger, openml-xgb and lassobench), there may be many other factors impacting performance and we do not have enough benchmarks to draw any firm conclusions around this. \\

\subsection{Limitations}

An obvious limitation to this work is the lack of mathematical relationship between source and target tasks. We use transfer learning in the hope that it will improve speed of convergence of BO. However, when applying this technique to a new benchmark, there is no guarantee that ensemble-based transfer learning will provide better performance of BO. Our experimental setup involves averaging over a number seeds and tasks and requires more work to find a way to apply with guarantees for a new problem where we do not have access to all historic tasks for cross validation.\\

 Another limitation arising from this is that the insights from our empirical study are not guaranteed to generalize to other benchmarks. While we have found some general trends that may relate to a wider set of benchmarks, more work is required to understand how specific pipeline components can be most effectively applied to particular characteristics of objective functions from different fields.\\

Finally, given that these experiments are all averaged over seed task combinations, there is no guarantee that it will work well for a particular target task. The experimental approach used here produces results based on averaging over all available tasks in a benchmark, and 15 seeds for each task. During development, we noted that performance can vary significantly for different tasks and seeds. In developing the approaches proposed here for a real optimisation problem we would like to know whether to use standard BO or ensemble-based transfer learning with BO on a new task for which we only have source historic datasets, and no access to source objective functions. However, without information about how various optimisation methods performed on historic tasks, it would be difficult to guess the best approach. This problem was the motivation for our developing an approach to historic data analysis. While this approach shows some promise, there are still development required to make this approach more reliable.\\

\subsection{Future Work}

In view of the limitations described, we recommend two specific areas of future work. These are firstly the development of a better strategy for detecting and handling bad transfer learning during optimisation. This is particularly challenging with small amounts of data. Secondly, further development of our proposed approach to analysing historic data is required to improve accuracy of predictions about effectiveness of transfer learning given access historic (source) datasets only and not historic (source) objective functions. Our method for using clustering may be contributing to inaccuracies in approximating location of minima, and further investigation is required.\\

\section{Acknowledgements}
The data used to construct our proposed benchmark, lassobench, obtained from the California Cooperative Oceanic Fisheries Investigations, is available at their website, \url{https://calcofi.org/}.


\appendix
\section{Important Extra Details}
\label{app1}

\subsection{Algorithm for Pre-learning Regularisation Hyperparameters}
\label{section:alpha_alg}

\begin{algorithm}[!htb]
\caption{PreLearnAlpha}
\label{algorithm:alpha}
\begin{algorithmic}[1]
    \Require $\mathcal{D}_{task} = \{\mathcal{D}_1,...,\mathcal{D}_N\}$ for total $N$ tasks, evaluation budget $\mathcal{T}$
    \State \textbf{Initialisation}
    \Statex a. Pre-train ensemble of GPs, $f_i \sim \mathcal{GP}(\mu_i, K_i)$ for $i \in 1,...,N$  
    \Statex b. Initialise empty alphas list
    \For{t = 1,2,...,N}
        \State Let t be pseudo-target task
        \State Training dataset is $\mathcal{D}_{pseudo-target} = \{\boldsymbol{x}_{pseudo-target,j}, y_{pseudo-target,j}\}_{j = 1}^{\mathcal{T}}$
        \State Compute best $\alpha$ for Lasso loss function using $\{f_i|i\neq t\}$ as source functions (see Eq \ref{equation:background_lasso_loss}) using cross validation on training dataset
        \State Append best $\alpha$ to alphas list
    \EndFor
    \Ensure Median $\alpha$ from alphas list    
\end{algorithmic}
\end{algorithm}

\subsection{More Benchmark Details}
\label{appendix:benchmarks}

In this section we provide details of variable name, range, and type for benchmarks. For those benchmarks 

\subsubsection{OpenML100 Further Details}
\label{appendix:openml100_details}

For openml-glmnet benchmark details see \cite[Appendix D, Table 9]{Feurer2022}.\\

For openml-svm benchmark details see \cite[Appendix D, Table 10]{Feurer2022}.\\

For openml-xgb benchmark details see \cite[Appendix D, Table 11]{Feurer2022}.\\

For openml-rpart benchmark details see \cite[Table 1]{Kuhn2018automatic}. \\

For openml-ranger benchmark details see \cite[Table 1]{Kuhn2018automatic}. \\

For implementation details of OpenML100 benchmarks see code accompanying \cite{Feurer2022} at \url{https://github.com/automl/transfer-hpo-framework/}. \\

\subsubsection{RandomForest (OpenML-CC18) Benchmark}
\label{appendix:randomforest_details}
The task IDs for OpenML-CC18 tasks included in our RandomForest benchmark are as follows: 1510, 40668, 41027, 40701, 1468, 1461, 4538, 32, 11, 1063, 40994, 14, 31, 1067, 1475, 1480, 4534, 1068, 22, 46, 469, 1501, 40975, 1494, 1497, 40982, 182, 40984, 1464, 50, 1462, 16, 54, 307, 40978, 23, 44, 40983.

Table \ref{tab:randomforest_bench} provides specific details of variables in Randomforest.

\begin{table}[t]
\centering
\begin{tabular}{l c r}
    \hline
    Name & Range & Variable Type \\[0.5ex] 
    \hline\hline
    n\_estimators & [1,200] & integer \\ 
    max\_depth & [1,200] & integer \\
    min\_samples\_split & [2,10] & integer \\
    min\_samples\_leaf & [1,5] & integer \\
    max\_leaf\_nodes & [100,3000] & integer \\
    min\_weight\_fraction\_leaf & [0.0,0.5] & continuous \\
    ccp\_alpha & [0.0,0.5] & continuous \\
    min\_impurity\_decrease & [0.0,0.5] & continuous \\
    criterion & ['gini', 'entropy', 'log\_loss'] & categorical \\
    max\_features & [None, 'sqrt', 'log2'] & categorical \\
\end{tabular}
\caption{Randomforest Benchmark Variables}\label{tab:randomforest_bench}
\end{table}

\subsubsection{LassoBench Benchmark}
\label{appendix:lassobench_details}
Table \ref{tab:lassobench_bench} provides specific details of variables in our LassoBench benchmark.

\begin{table}[t]
\centering
\begin{tabular}{l c r}
    \hline
    Name & Range & Variable Type \\[0.5ex] 
    \hline\hline
    $\alpha_1$ & [-1,1] & continuous \\ 
    $\alpha_2$ & [-1,1] & continuous \\
    $\alpha_3$ & [-1,1] & continuous \\
    $\alpha_4$ & [-1,1] & continuous \\
    $\alpha_5$ & [-1,1] & continuous \\
    $\alpha_6$ & [-1,1] & continuous \\
    $\alpha_7$ & [-1,1] & continuous \\
    $\alpha_8$ & [-1,1] & continuous \\
    $\alpha_9$ & [-1,1] & continuous \\
    $\alpha_{10}$ & [-1,1] & continuous \\
\end{tabular}

\caption{Lassobench Benchmark Variables}\label{tab:lassobench_bench}
\end{table}

\subsubsection{Cartpole Benchmark}
\label{appendix:cartpole_details}

The cartpole benchmark, contributed as part of this work, was adapted from the setup descibed in \cite{Marco2017virtual} which used Bayesian optimisation over multiple information sources (simulation and physical model), to the transfer learning with BO context. We construct this benchmark as an example (an extension of a synthetic benchmark), rather than claiming it to be inately useful in the context of control. While \cite{Marco2017virtual} uses multi-task learning scenario to swap between evaluations of real and simulated cartpole, this paper uses the parameters provided by the Quanser User Manual, accessed via \cite{QuanserSIP, QuanserIP01}, for the real cartpole in \cite{Marco2017virtual}, to simulate a range of similar cartpoles with LQR controller. Parameters for similar cartpoles are sampled from a uniform distribution (see \cite{Tighineanu2022} for a similar idea with synthetic functions) with ranges as shown in Table \ref{tab:cartpole_params}. The loss function used as the black-box objective function, from \cite{Marco2017virtual}, is

\begin{equation}
\label{equation:cartpole_cost}
    J = \frac{1}{K}\sum_{K=0}^{K-1} s_k^2 + \psi_k^2 + \dot{s_k^2} + 0.1\dot{\psi_k^2} + 10^{-5}u_k^2.
\end{equation}

In this equation, $J$ is used to denote cost, $s$ is the position of the cart, $\psi$ is the angle of the pole and $u$ is the control variable (voltage).Algorithm \ref{algorithm:cartpole} provides a rough outline for this benchmark, and Table \ref{tab:cartpole_bench} describes the BO variables. The variables, $\theta_1$ and $\theta_2$ are tuning parameters for the $Q=diag(10^{\theta_1}, 1, 1 , 0.1)$ and $R=10^{-\theta_2}$ LQR weights. \\

\begin{algorithm}
\caption{Cartpole}
\label{algorithm:cartpole}
\begin{algorithmic}[1]
    \Require cart mass, pole mass, pole length, cart friction, pole friction, evaluation budget $\mathcal{T}$, number simulation time steps $K$
    \State \textbf{Initialisation}
    \State Compute A,B (linearised version of dynamics)   
    \State Initialise empty evaluations list, $evals$
    \State Initialise Q,R (LQR weights)
    \State Initialise simulation state vector, $x$
    \For{$i = 1,2,...,\mathcal{T}$}
        \State Compute $\mathcal{K} = \Call{lqr}{A,B,Q,R}$*
        \State $loss \gets 0$
        \For{$k = 1,2,...,K$}
            \State Update $u = \mathcal{K}*x$
            \State $loss += \Call{ComputeLoss}{x,u}$ (see Equation \ref{equation:cartpole_cost})
            \State Update $x = \Call{SimulationStep}{x,u}$**
        \EndFor
        \State $evals[i] = loss$
        \State Update BO surrogate model (standard BO or transfer learning BO)
        \State Update Q,R by optimising acquisition function
    \EndFor
    \Ensure $evals$
\end{algorithmic}
\end{algorithm}

* We used \cite{Fullerpython-control2021} control package for this.
** Simulation used cartpole equations derived based on description in \cite[Chapter 3]{Tedrake2024}

\begin{table}[t]
\centering
\begin{tabular}{l c r}
    \hline
    Name & Range & Variable Type \\[0.5ex] 
    \hline\hline
    $\theta_1$ & [-3,2] & continuous \\ 
    $\theta_2$ & [1,5] & continuous \\
\end{tabular}
\caption{Cartpole Benchmark Variables}\label{tab:cartpole_bench}
\end{table}

\begin{table}[t]
\centering
\begin{tabular}{l c r}
    \hline
    Name & Range \\[0.5ex] 
    \hline\hline
    cart mass & [0.1,0.5] \\ 
    pole mass & [0.01,0.25] \\
    pole length & [0.25,0.75] \\
    cart friction & [0.0001, 0.001] \\
    pole friction & [0.001, 0.01] \\
\end{tabular}

\caption{Cartpole Simulation Parameters}\label{tab:cartpole_params}
\end{table}

\subsection{Some Additional Plots for Main Results}

\subsubsection{Additional Normalised Regret Plots for Handling Bad Transfer Learning}
\label{appendix:norm_regret_bad_tl}

The section contains extra plots for comparing methods with and without a strategy for handling poor transfer learning. RiGPE with positive constraint, RiGPE without positive constraint, and TSTR are included here. See Section \ref{section:main_results_bad_transfer_learning} for related plots comparing weighting strategies inclusive of strategy for handling poor transfer learning.\\

\begin{figure}
\includegraphics[width=1.0\textwidth]{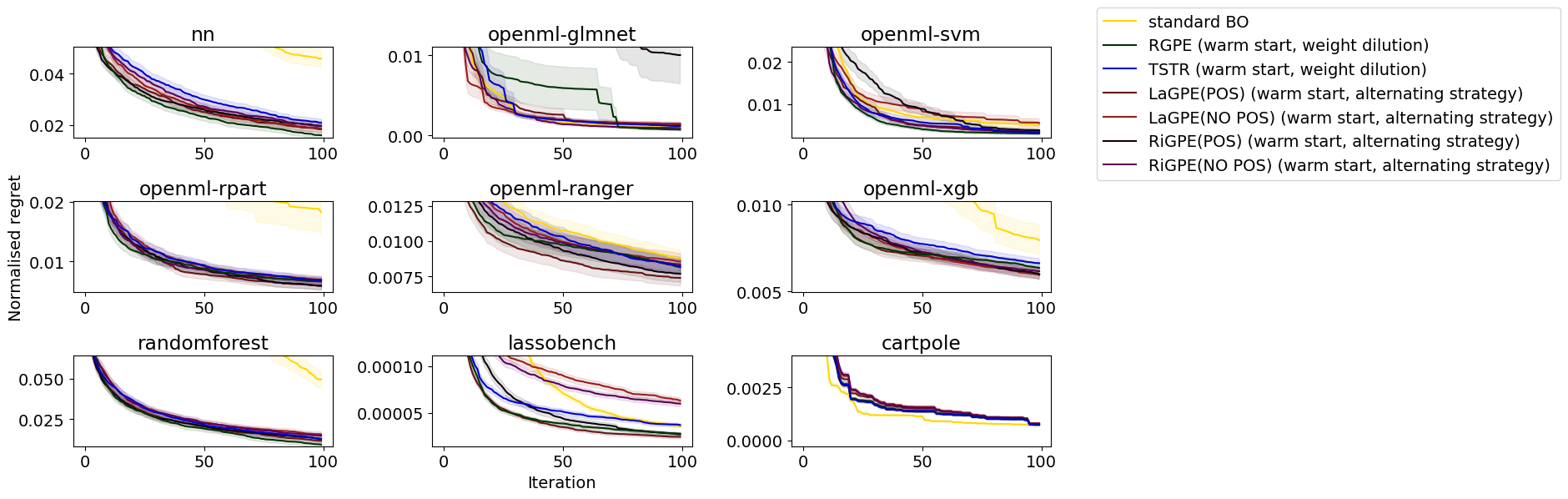}
    \caption{Plots comparing different weighting strategies using full pipeline}\label{fig:full_plots}
\end{figure}

\begin{figure}
\includegraphics[width=1.0\textwidth]{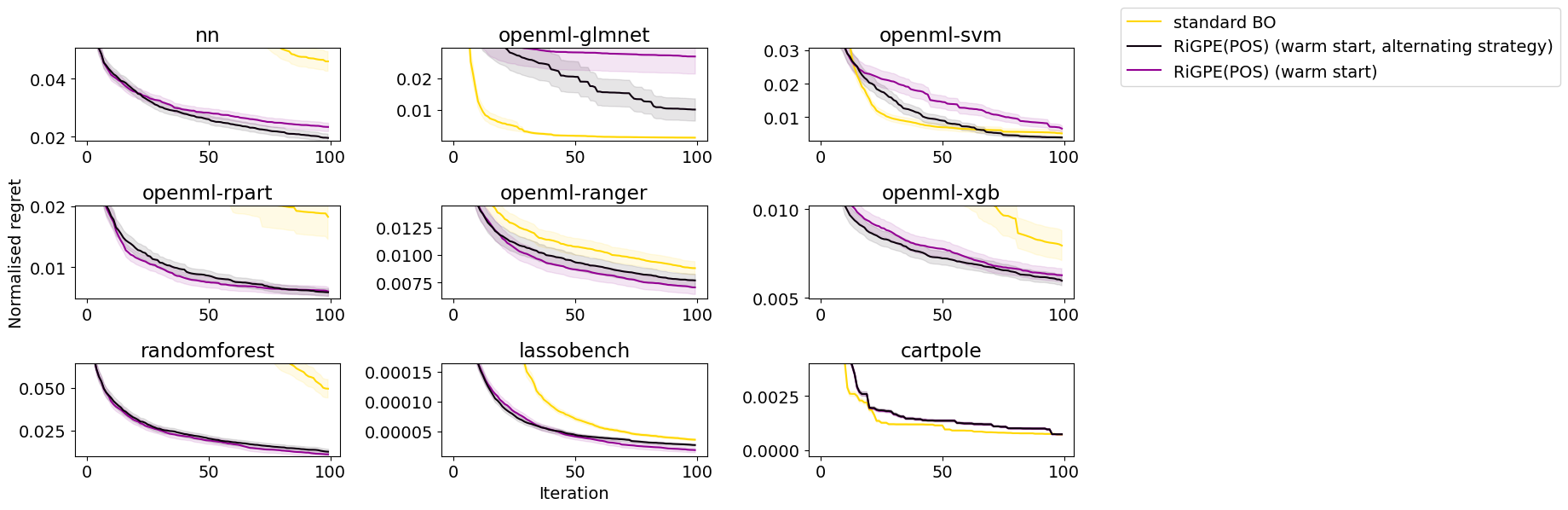}
    \caption{RiGPE (with positive constraint on the weights) plots with and without strategy to handle bad transfer learning}\label{fig:rigpe_bad_tl}
\end{figure}

\begin{figure}
\includegraphics[width=1.0\textwidth]{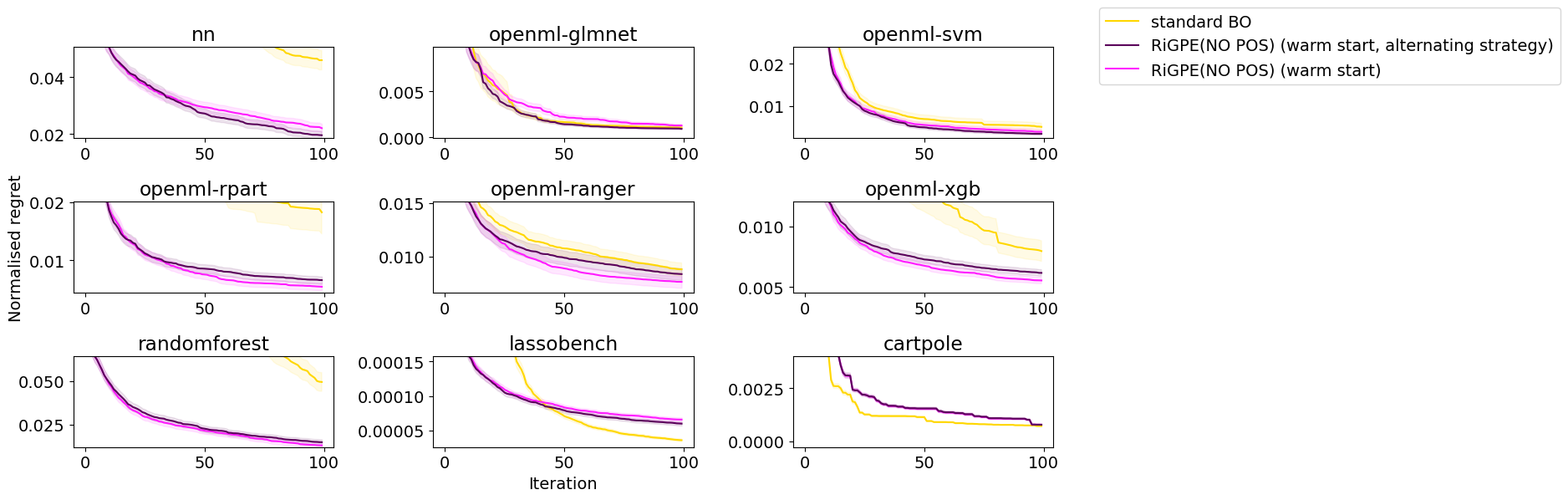}
    \caption{RiGPE (no positive constraint on weights) plots with and without strategy to handle bad transfer learning}\label{fig:rigpe_np_bad_tl}
\end{figure}

\begin{figure}
\includegraphics[width=1.0\textwidth]{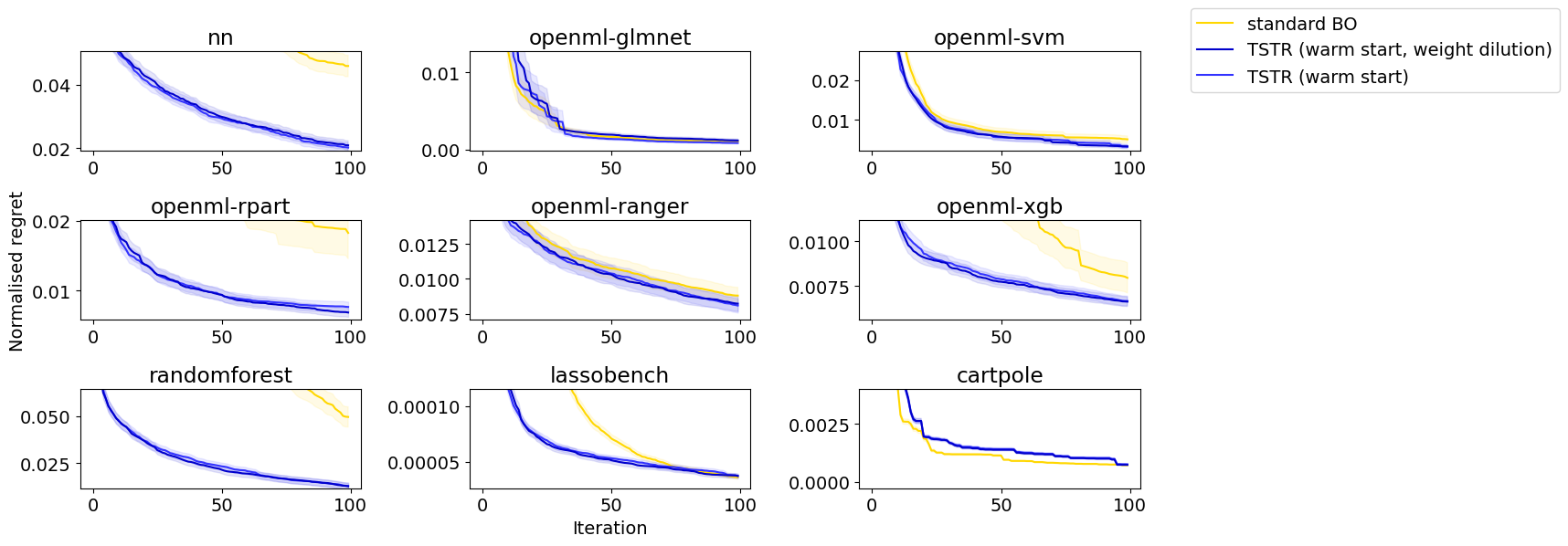}
    \caption{TSTR plots with and without strategy to handle bad transfer learning}\label{fig:TSTR_bad_tl}
\end{figure}

\subsubsection{Scatter Plots for Analysis of Historic Data}
\label{appendix:scatter_plots_historic_data}

\begin{figure}
    \centering
    \subfloat[]{
        \includegraphics[scale=0.11]{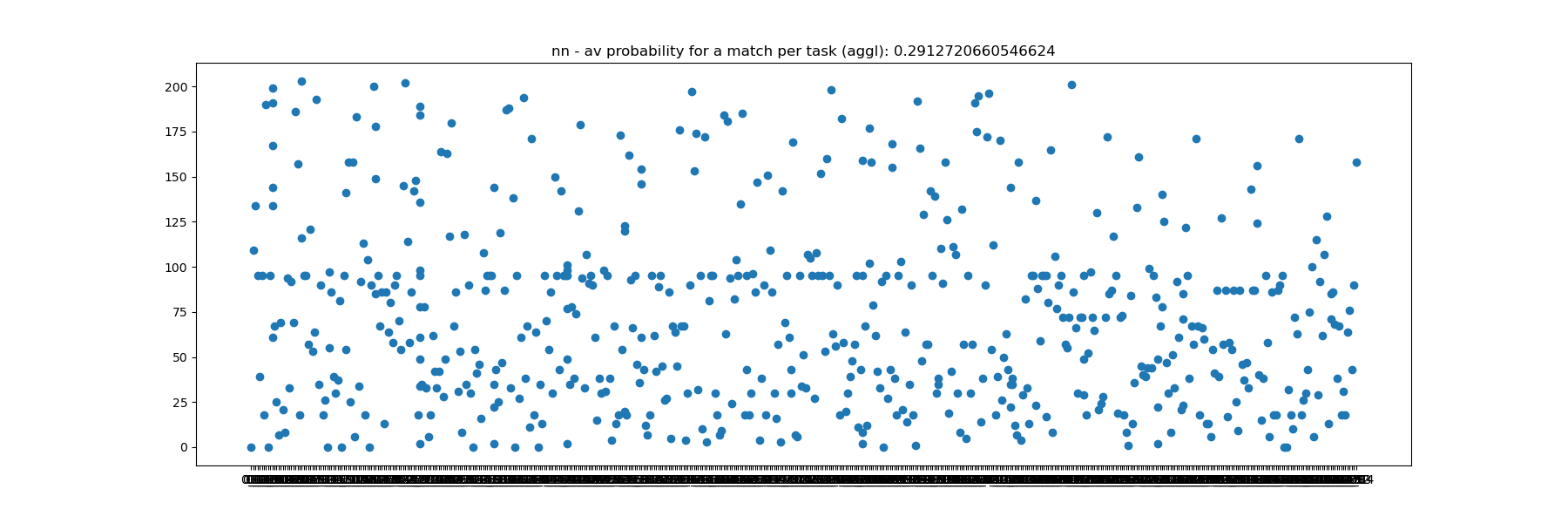}\label{nn_agg}
    }\hspace{0.4cm}
    \subfloat[]{
        \includegraphics[scale=0.11]{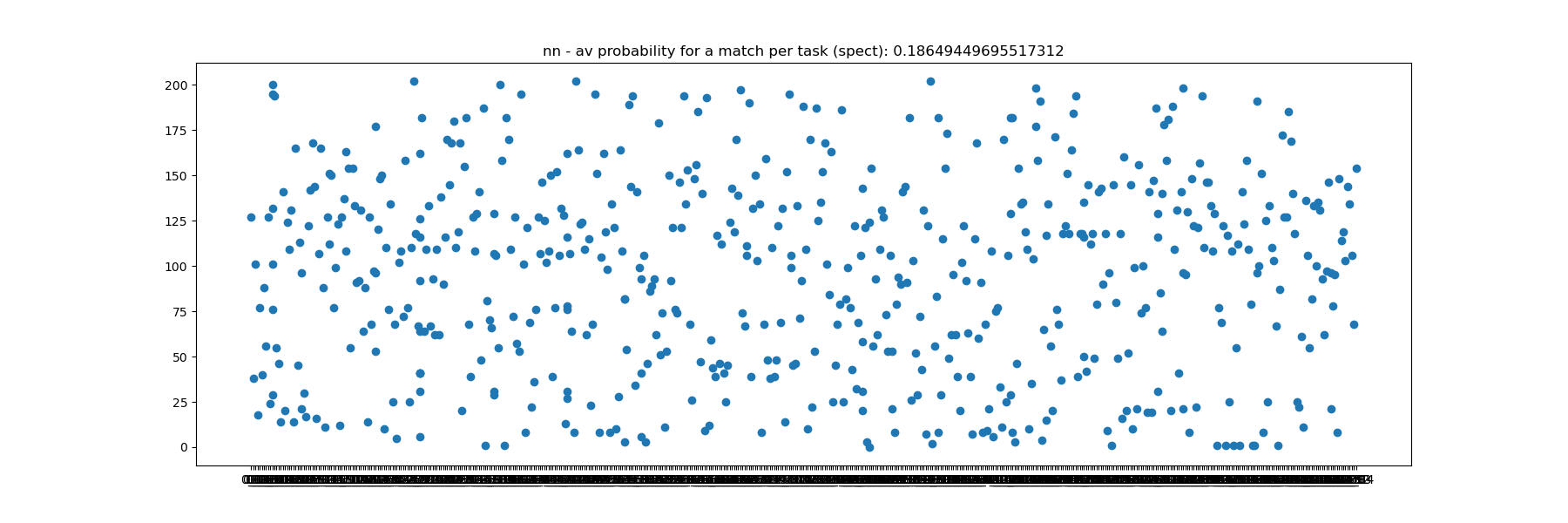}\label{nn_spect}
    }\hspace{0.4cm}\\
        \subfloat[]{
        \includegraphics[scale=0.11]{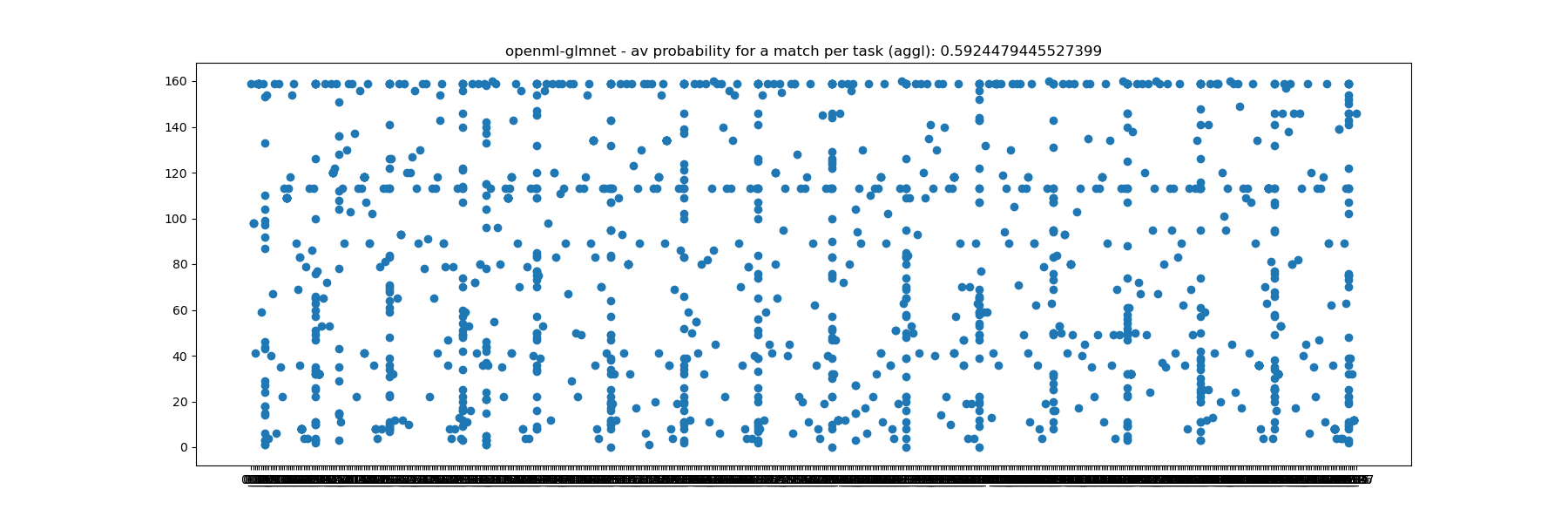}\label{glmnet_agg}
    }\hspace{0.4cm}
    \subfloat[]{
        \includegraphics[scale=0.11]{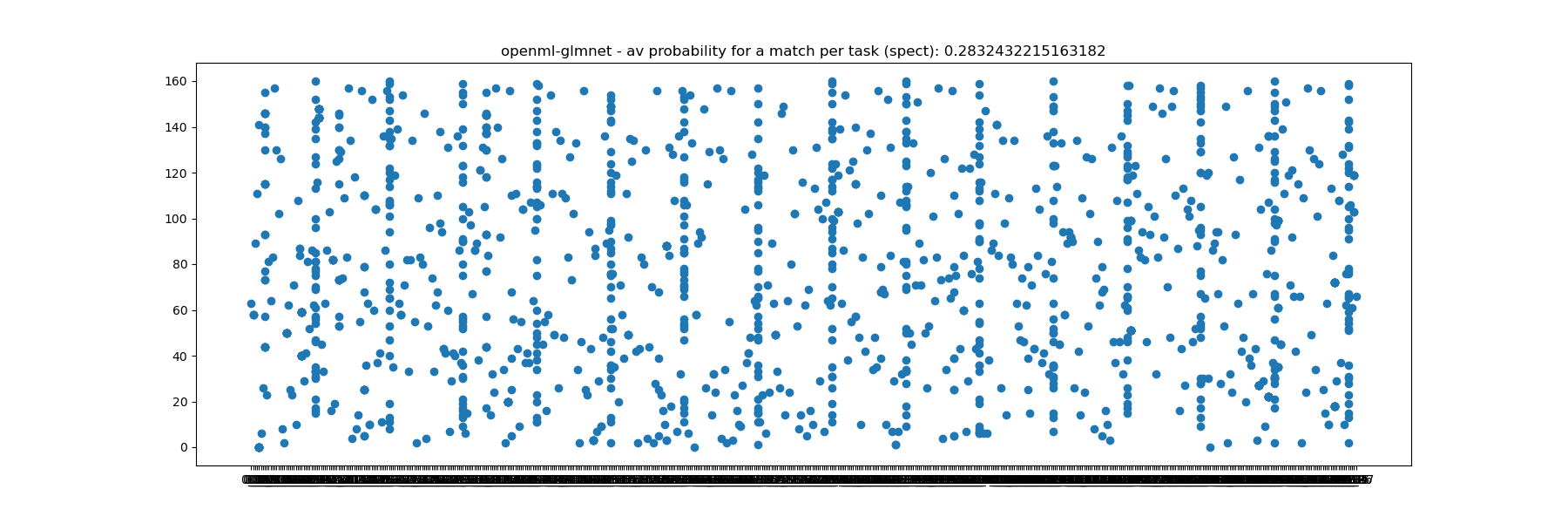}\label{glmnet_spect}
    }\hspace{0.4cm}\\
        \subfloat[]{
        \includegraphics[scale=0.11]{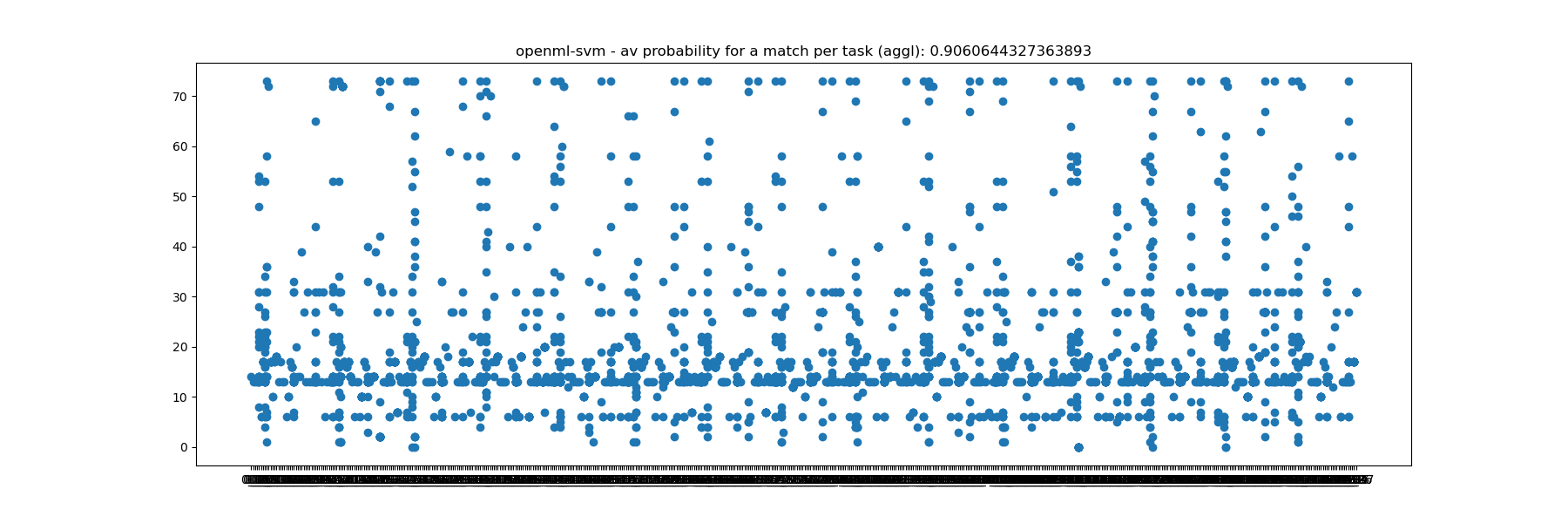}\label{svm_agg}
    }\hspace{0.4cm}
    \subfloat[]{
        \includegraphics[scale=0.11]{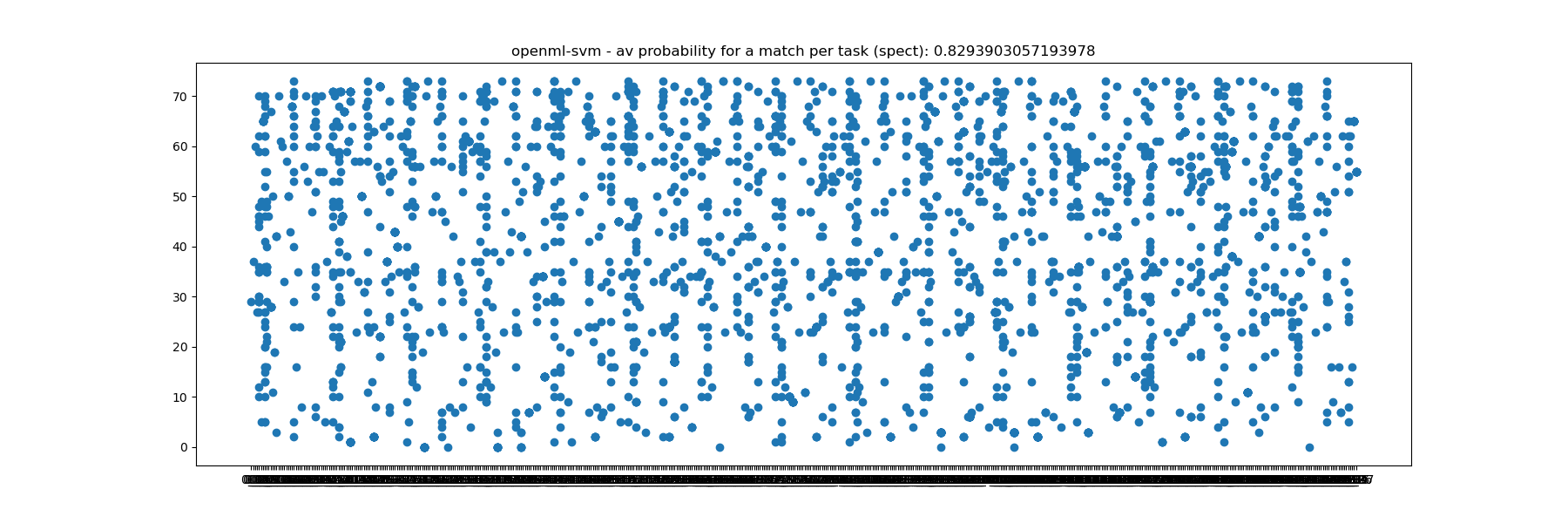}\label{svm_spect}
    }\hspace{0.4cm}\\
        \subfloat[]{
        \includegraphics[scale=0.11]{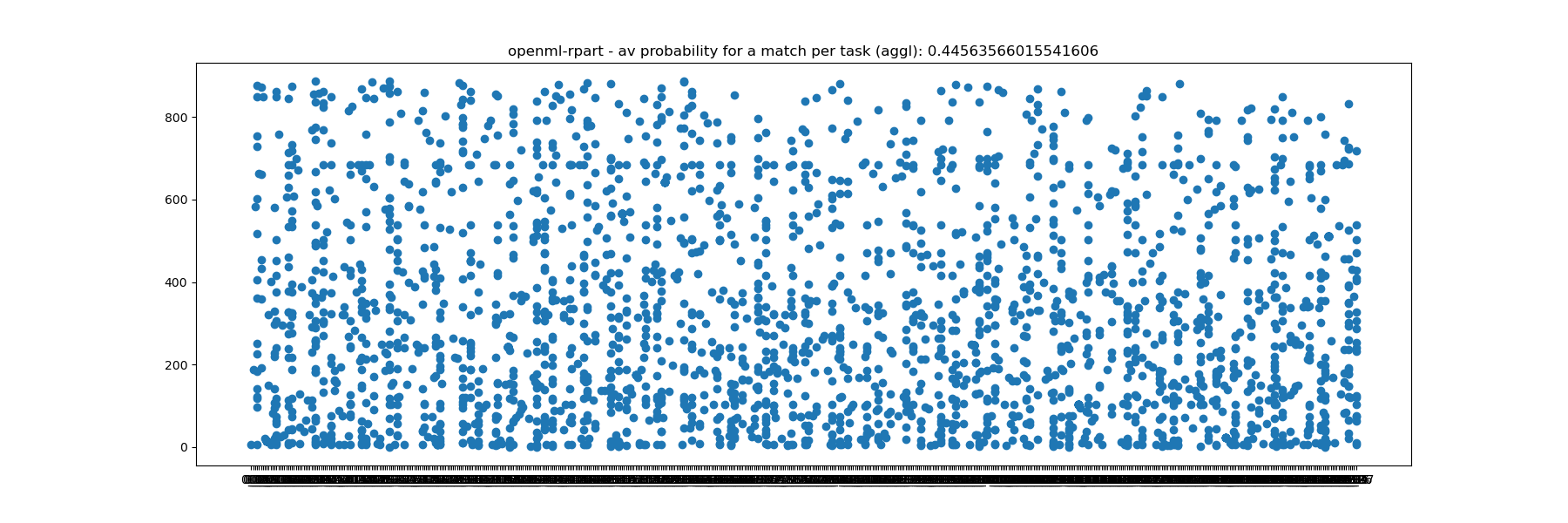}\label{rpart_agg}
    }\hspace{0.4cm}
    \subfloat[]{
        \includegraphics[scale=0.11]{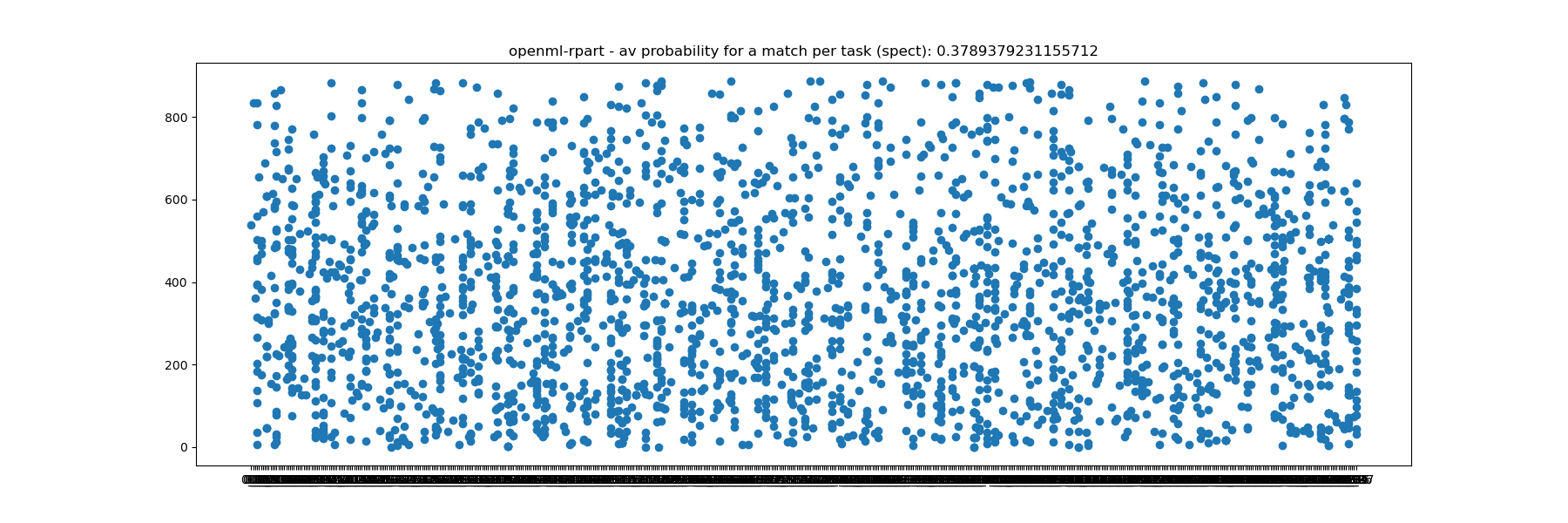}\label{rpart_spect}
    }\hspace{0.4cm}\\
        \subfloat[]{
        \includegraphics[scale=0.11]{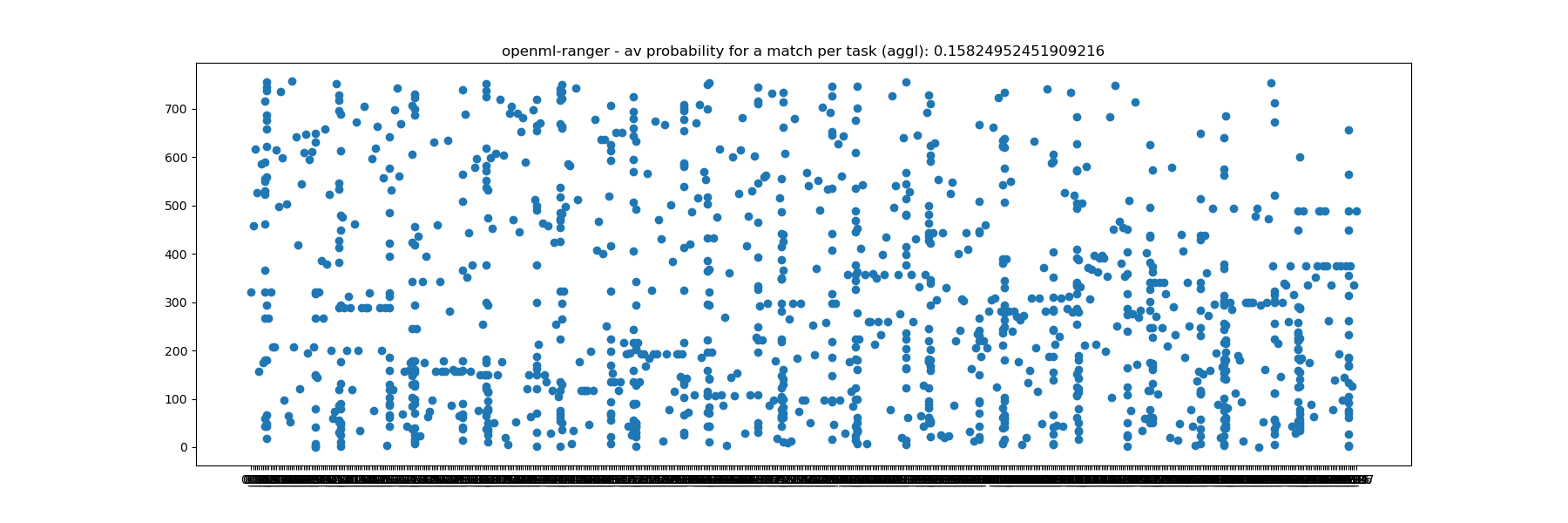}\label{ranger_agg}
    }\hspace{0.4cm}
    \subfloat[]{
        \includegraphics[scale=0.11]{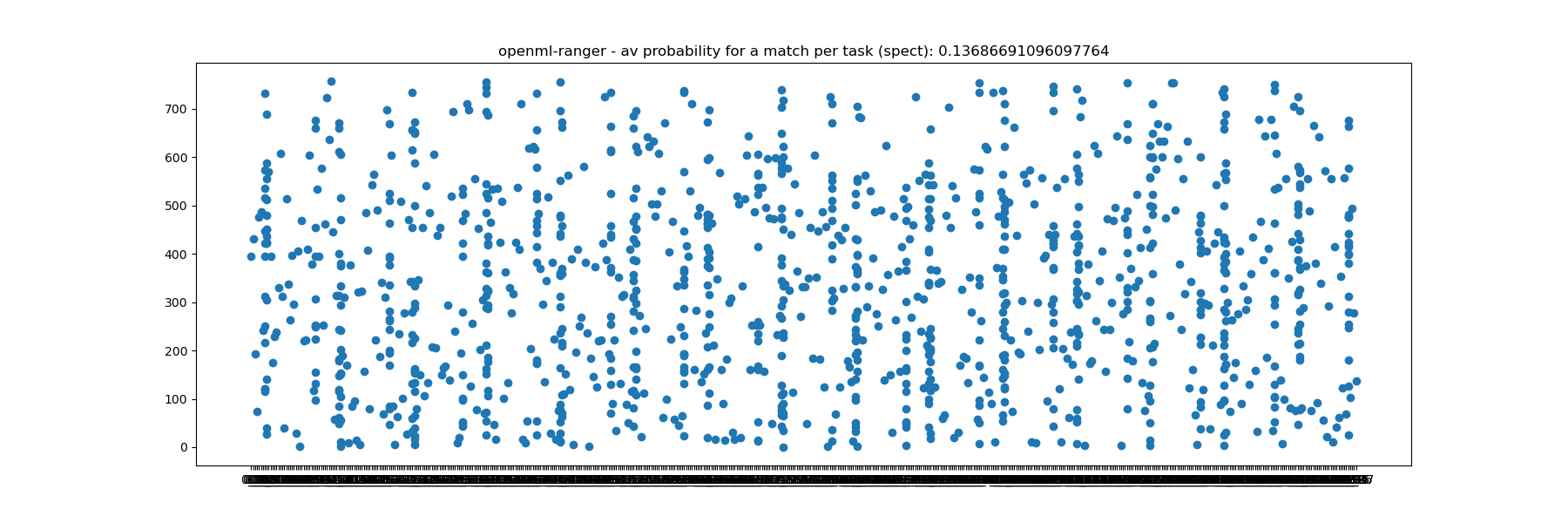}\label{ranger_spect}
    }\hspace{0.4cm}\\
        \subfloat[]{
        \includegraphics[scale=0.11]{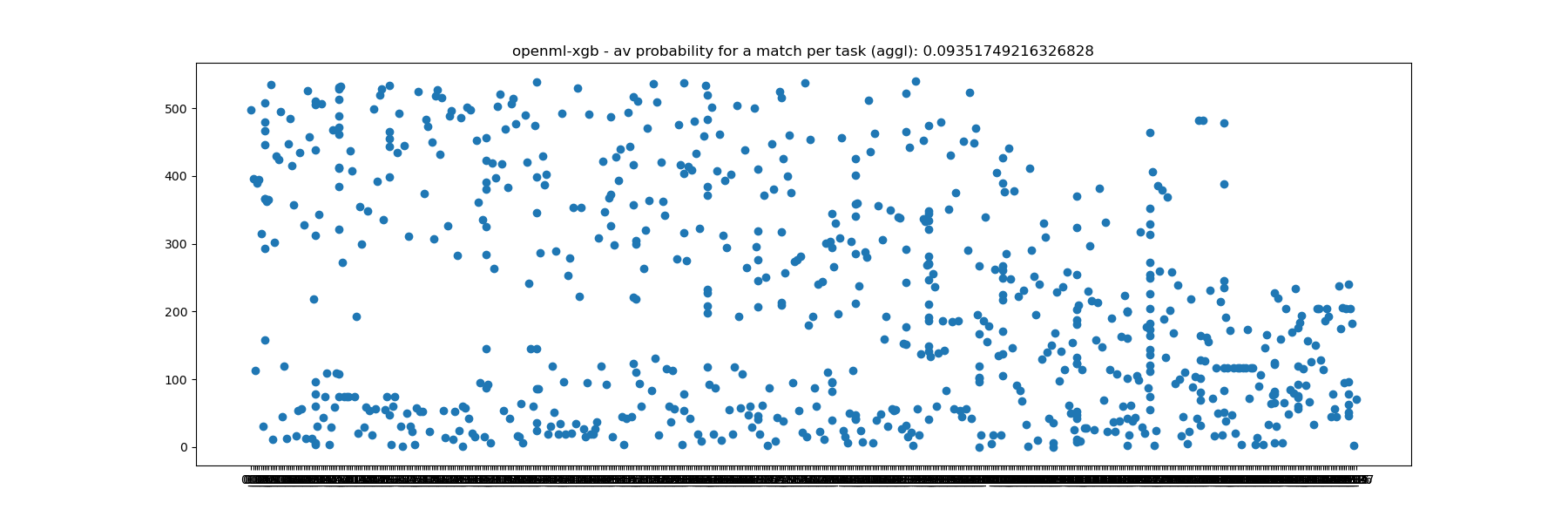}\label{xgb_agg}
    }\hspace{0.4cm}
    \subfloat[]{
        \includegraphics[scale=0.11]{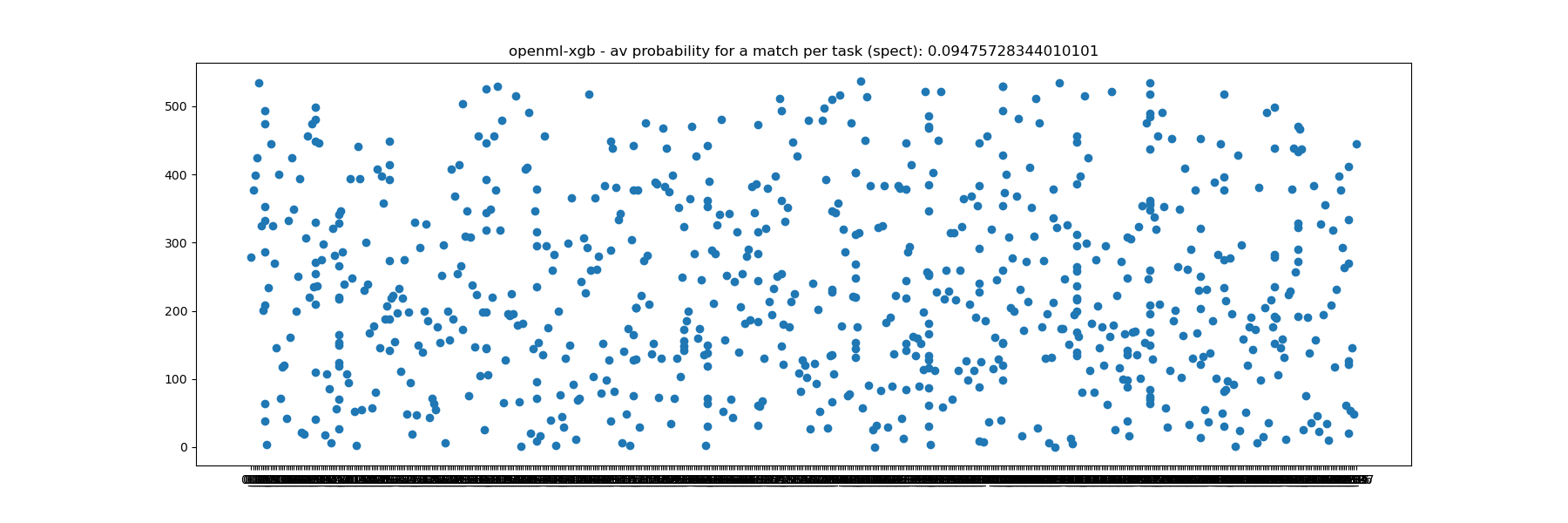}\label{xgb_spect}
    }\hspace{0.4cm}\\
        \subfloat[]{
        \includegraphics[scale=0.11]{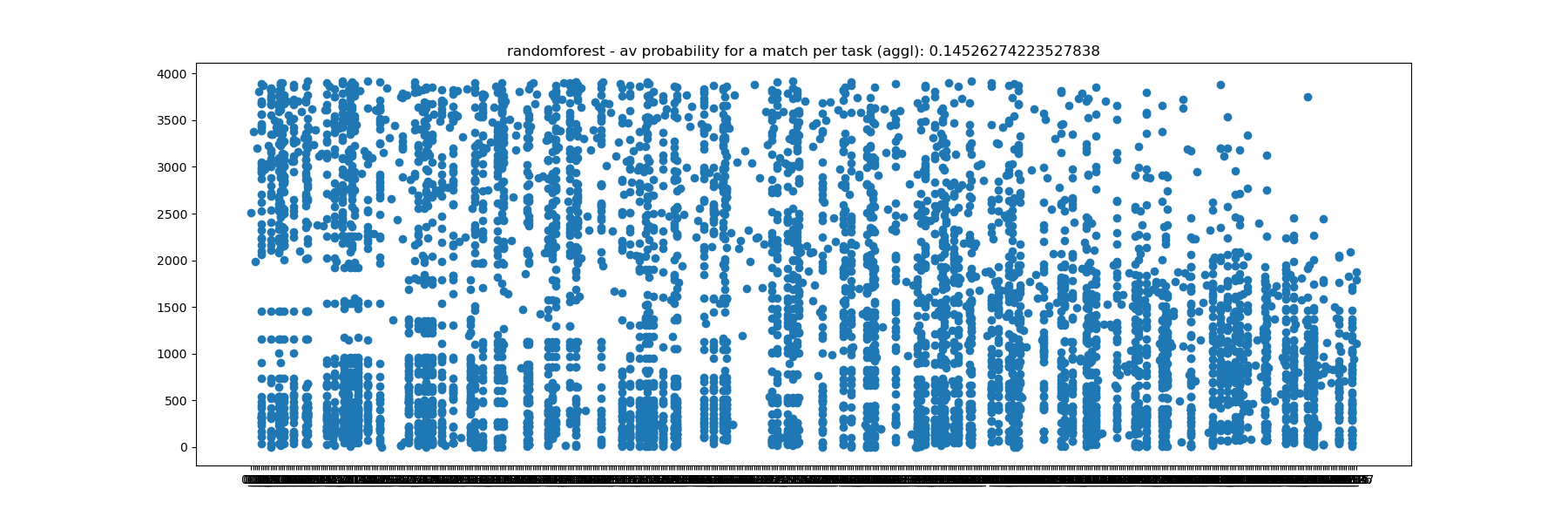}\label{randomforest_agg}
    }\hspace{0.4cm}
    \subfloat[]{
        \includegraphics[scale=0.11]{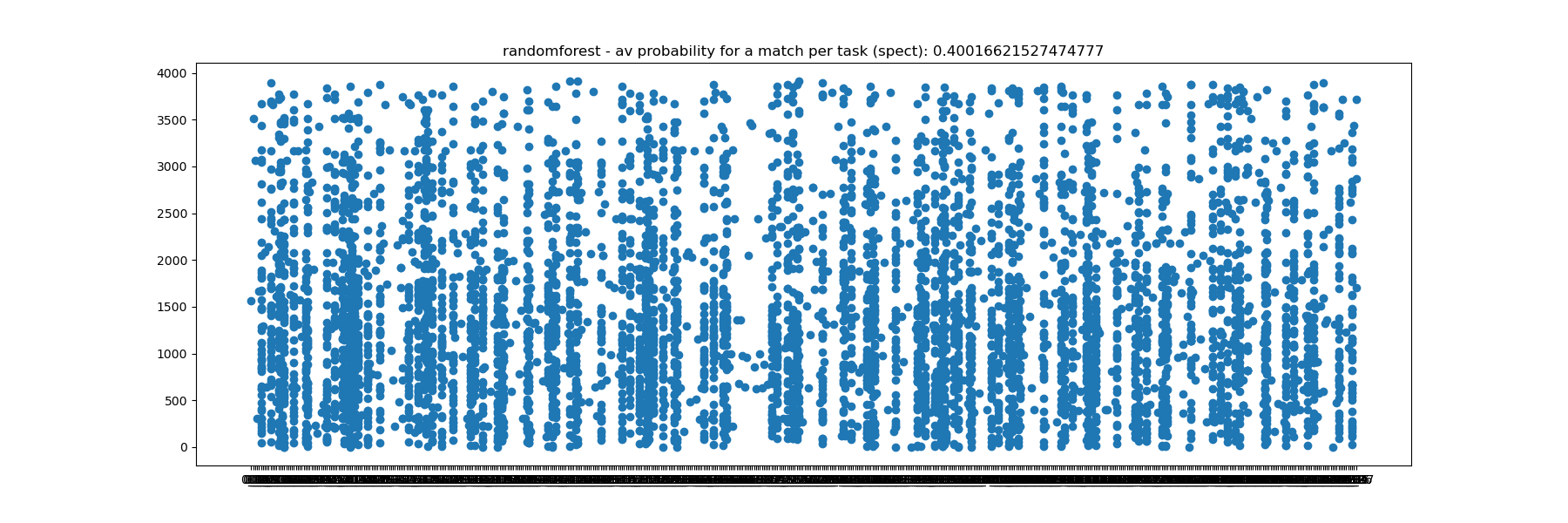}\label{randomforest_spect}
    }\hspace{0.4cm}\\
        \subfloat[]{
        \includegraphics[scale=0.11]{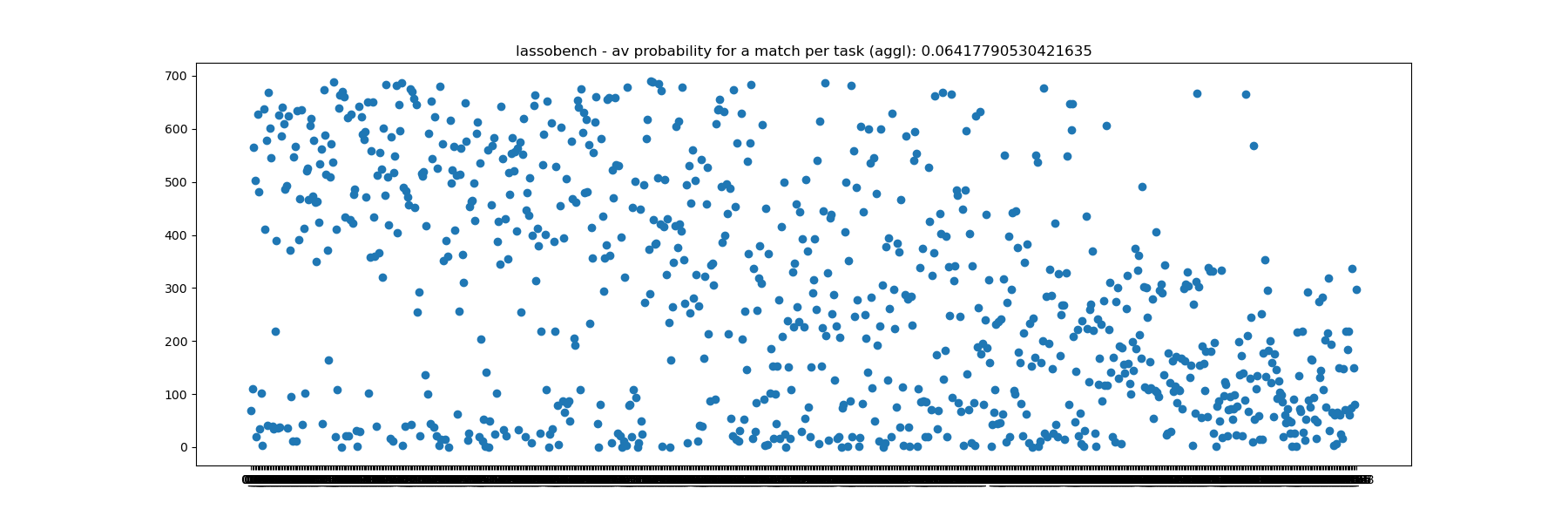}\label{lassobench_agg}
    }\hspace{0.4cm}
    \subfloat[]{
        \includegraphics[scale=0.11]{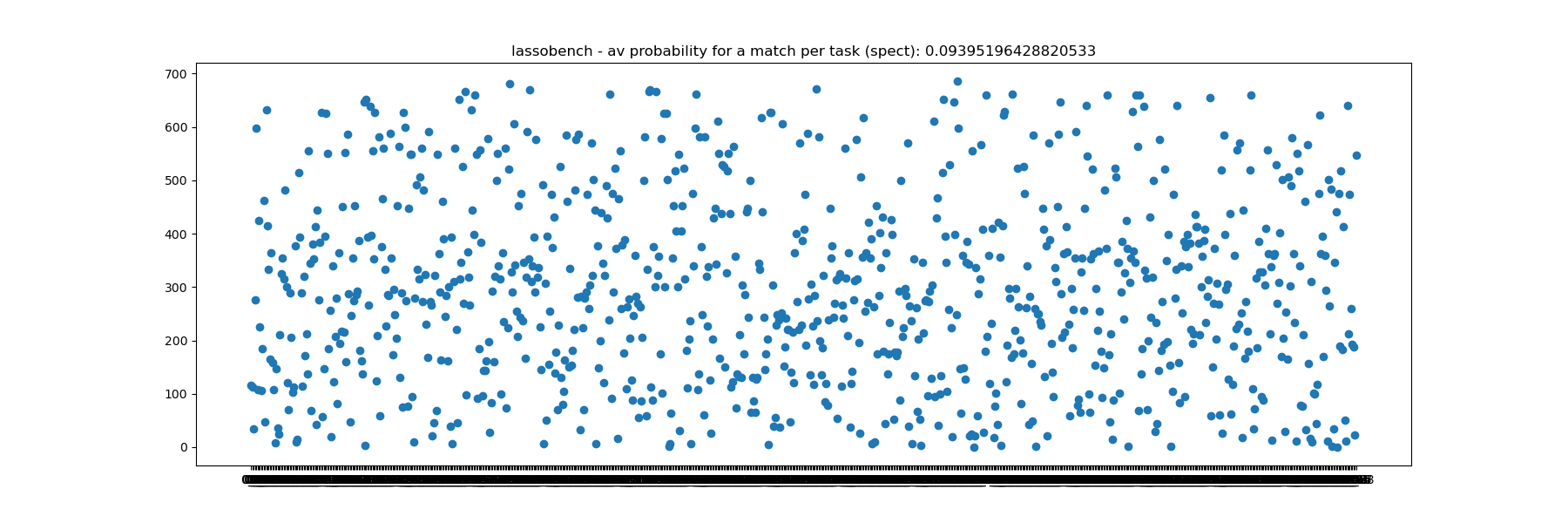}\label{lassobench_spect}
    }\hspace{0.4cm}\\
        \subfloat[]{
        \includegraphics[scale=0.11]{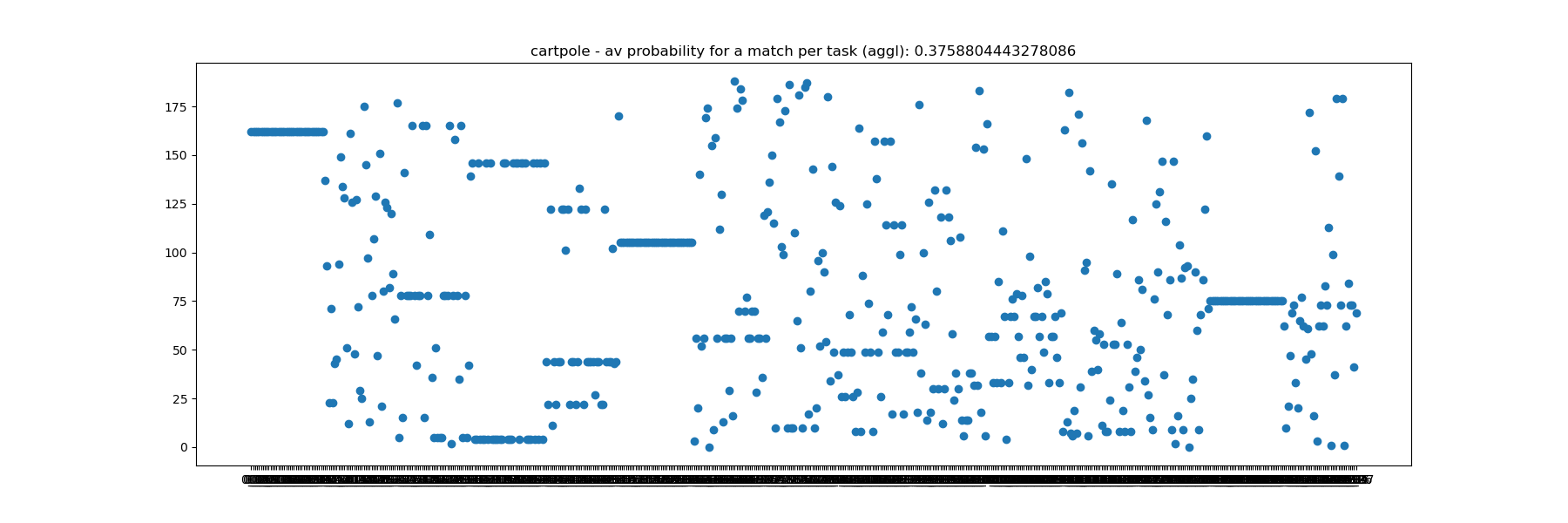}\label{cartpole_agg}
    }\hspace{0.4cm}
    \subfloat[]{
        \includegraphics[scale=0.11]{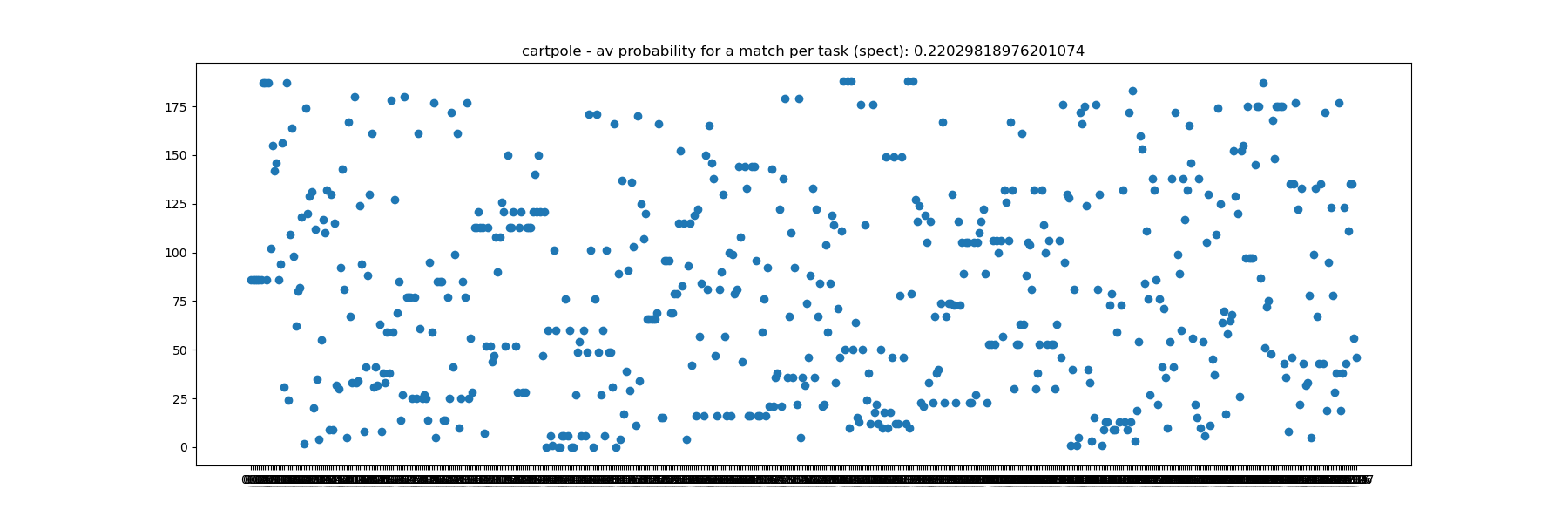}\label{cartpole_spect}
    }\hspace{0.4cm}\\
    \caption{Scatter plots for Agglomerative Clustering and Spectral Clustering. x-axis is seed-task and y-axis is cluster number.}\label{fig:cluster_scatter_plots}
\end{figure}

Due to the experimental nature of clustering \cite{Xiong2018clustering}, and the absence of suitable labels for validating the use of agglomerative clustering, we also include spectral clustering, with number of clusters set to be the same as that discovered by agglomerative clustering with distance threshold of $0.02$ and complete linkage. Scatter plots for each benchmark with agglomerative clustering on the left and spectral clustering on the right are shown in Figure \ref{fig:cluster_scatter_plots}. In these plots, the x-axis is seed-task, and y-axis is cluster number. While numbering of clusters will not be in identical order using the two clustering algorithms, the overall clustering patterns can be seen to be similar for most benchmarks.\\

\bibliographystyle{abbrv} 
\bibliography{bibliography}

\end{document}